\documentclass[acmtog]{acmart}
\acmSubmissionID{171}

\usepackage{booktabs} 

\usepackage[ruled]{algorithm2e} 

\SetAlFnt{\small}
\SetAlCapFnt{\small}
\SetAlCapNameFnt{\small}
\SetAlCapHSkip{0pt}

\setcopyright{acmcopyright}
\acmJournal{TOG}
\acmYear{2022}
\acmVolume{41}
\acmNumber{6}
\acmArticle{231}
\acmMonth{12} 
\acmDOI{10.1145/3550454.3555442}

\usepackage{microtype}
\usepackage{caption}
\usepackage{subcaption}
\usepackage{xcolor}

\newcommand{\YY}[1]{#1}

\newcommand{\etal}{\textit{et al.}}

\newcommand{\sbar}[1]{\bar{#1}}
\newcommand{\dbar}[1]{\widetilde{#1}}
\newcommand{\lens}[1]{{#1}_l}

\begin{document}
\title{Learning to Relight Portrait Images via a Virtual Light Stage and Synthetic-to-Real Adaptation}

\author{Yu-Ying Yeh}
\affiliation{%
 \institution{University of California, San Diego}
 \country{USA}
}
\affiliation{%
 \institution{NVIDIA}
 \country{USA}
}
\authornote{Work done during an internship at NVIDIA, USA}
\email{yuyeh@eng.ucsd.edu}
\author{Koki Nagano}
\affiliation{%
 \institution{NVIDIA}
 \country{USA}
}
\email{koki.nagano0219@gmail.com}
\author{Sameh Khamis}
\affiliation{%
 \institution{NVIDIA}
 \country{USA}
}
\email{sameh@umiacs.umd.edu}
\author{Jan Kautz}
\affiliation{%
 \institution{NVIDIA}
 \country{USA}
}
\email{jkautz@nvidia.com}
\author{Ming-Yu Liu}
\affiliation{%
 \institution{NVIDIA}
 \country{USA}
}
\email{mingyul@nvidia.com}
\author{Ting-Chun Wang}
\affiliation{%
 \institution{NVIDIA}
 \country{USA}
}
\begin{abstract}
Given a portrait image of a person and an environment map of the target lighting, portrait relighting aims to re-illuminate the person in the image as if the person appeared in an environment with the target lighting. To achieve high-quality results, recent methods rely on deep learning. An effective approach is to supervise the training of deep neural networks with a high-fidelity dataset of desired input--output pairs, captured with a light stage. However, acquiring such data requires an expensive special capture rig and time-consuming efforts, limiting access to only a few resourceful laboratories. To address the limitation, we propose a new approach that can perform on par with the state-of-the-art (SOTA) relighting methods without requiring a light stage. Our approach is based on the realization that a successful relighting of a portrait image depends on two conditions. First, the method needs to mimic the behaviors of physically-based relighting. Second, the output has to be photorealistic. To meet the first condition, we propose to train the relighting network with training data generated by a virtual light stage that performs physically-based rendering on various 3D synthetic humans under different environment maps. To meet the second condition, we develop a novel synthetic-to-real approach to bring photorealism to the relighting network output. In addition to achieving SOTA results, our approach offers several advantages over the prior methods, including controllable glares on glasses and more temporally-consistent results for relighting videos.
\end{abstract}



\begin{teaserfigure}
\centering
    \includegraphics[width=1.0\textwidth]{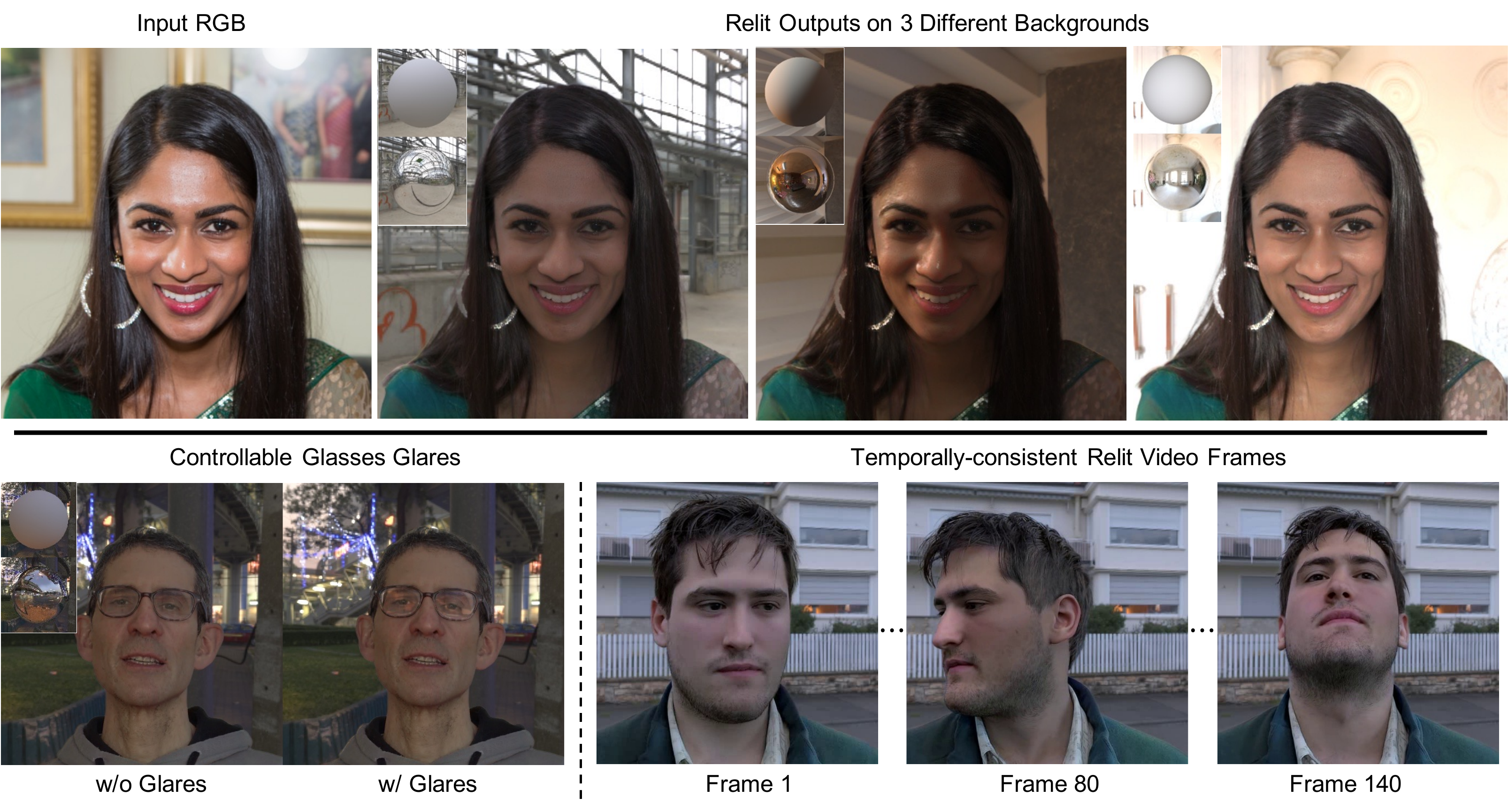}
    \vspace{-0.8cm}
	\caption{
	Top: Given an input portrait image and a target high dynamic range environment map (rendered as a diffuse ball and a mirror ball in the inset), our relighting method generates relit results under different illuminations with high photorealism. This is achieved by learning relighting physics from a synthetic dataset rendered using a virtual light stage, as well as adapting the learned model with a real-world dataset to close the domain gap.
	Bottom: our framework also enables new features, such as controlling the glares on glasses \YY{(diffuse and mirror balls shown in the inset)}, as well as synthesizing more temporally-consistent relit videos. Please refer to the supplementary video for more examples. Project page: \url{https://research.nvidia.com/labs/dir/lumos/}
	}
	\label{fig:teaser}
\end{teaserfigure}

\maketitle

\section{Introduction}

Given a photograph of a person, portrait relighting aims to re-illuminate the person as if they appeared in a new lighting environment. While this has been long used in professional settings, the growing interests in smartphone photography and video conferencing tools create a need for casually enhancing~\cite{nagano2019deep,zhang2020portrait} and re-illuminating a person from an in-the-wild image or video~\cite{pandey2021total}. Relighting a person as if they really belong to the scene requires knowledge of the 3D scene geometry and materials to simulate complex lighting interactions, including global illumination. Therefore, portrait relighting from a 2D in-the-wild portrait without any prior knowledge is extremely challenging. Due to the under-constrained nature of the problem, previous work relied on 3D face priors~\cite{shu2017portrait,Dib2021raytracing}, intrinsic images and shape from shading~\cite{BarronTPAMI2015, sengupta2018sfsnet, Kanamori2018fullbody, hou2022face}, or style transfer~\cite{shih2014transfer}. However, these models often fail to handle complex skin surfaces and subsurface reflectance, wide varieties of clothes or hairstyles, or complex inter-reflections between eyeglasses and other accessories. Capture-based data-driven techniques~\cite{debevec2000acquiring,wenger2005performance,weyrich2006analysis} have been proposed to achieve realistic relighting of human faces, but every subject needs to be recorded by a complex capture rig, limiting their applicability. Recently, the most successful methods for in-the-wild portrait relighting are based on deep learning. They train deep neural networks via a supervised framework using high-quality ground truth data captured by a light stage.

A light stage~\cite{debevec2000acquiring,debevec2012light,guo2019relightables} is a spherical lighting rig equipped with hundreds or thousands of programmable inward-pointing light sources evenly distributed on the sphere. As each light source provides a unique lighting direction, it is useful to capture an object illuminated by one-light-at-a-time (OLAT). The OLAT images can be used as point-based lighting bases to realistically relight faces under novel environments via image-based  relighting~\cite{debevec2000acquiring,wenger2005performance}. Several prior portrait relighting methods~\cite{sun2019single,pandey2021total,zhang2021neuralvideo} have used this technique to create the dataset necessary for training their portrait relighting networks. Each sample in the dataset consists of pairs of relit portrait images and their corresponding target environment maps. The network is then trained to predict a relit image given an input portrait and a target environment map. While valuable for ground truth computation, constructing a light stage is an expensive task. Moreover, building a light stage is only the first step. One still needs to find many subjects to sit still inside the light stage for data collection. Capturing and processing the resulting OLAT scans also requires time-consuming and labor-intensive efforts.

To challenge the conventional wisdom of the light stage data being the necessity of achieving high-quality portrait relighting, we develop a new approach that can achieve comparable results with the state-of-the-art methods without relying on a physical light stage. Our approach is based on the key observation that we can decouple the relighting problem into two: 1) learning \YY{to approximate the physically-based lighting behaviors from synthetic data produced by a path tracer} and 2) learning to synthesize photorealistic portrait images.

\begin{figure}[t]
\centering
\includegraphics[width=\linewidth]{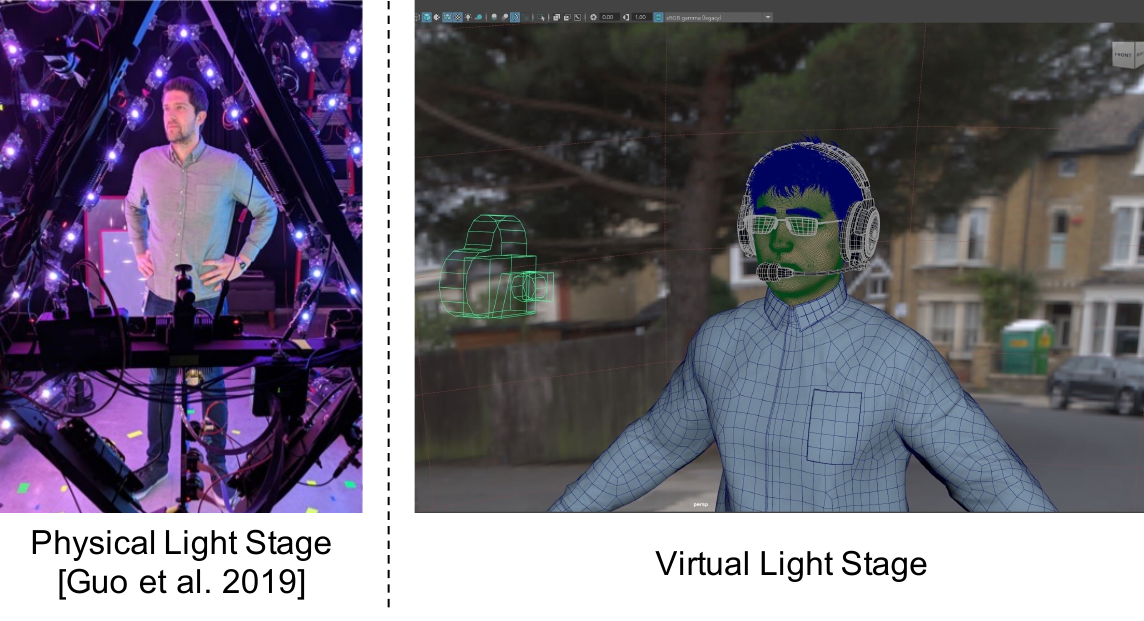}
    \vspace{-0.7cm}
  \caption{Illustration of a virtual light stage versus a physical light stage. Our virtual light stage setup renders high dynamic range images from a virtual camera. We use HDR environment maps as skydome lighting and randomly select one subject from our 3D scan collections paired with randomly selected hairstyles, clothes, and accessories. }
   \vspace{-0.3cm}
\label{fig:virtual_light_stage}
\end{figure}

We tackle the first part by leveraging advancement in physically-based rendering~\cite{pharr2016physically}. Specifically, we build a virtual light stage to re-illuminate pre-captured photogrammetry face scans to construct the training dataset using a physically-based renderer, as shown in~Figure~\ref{fig:virtual_light_stage}. Using the virtual light stage provides several key advantages over the physical one. First, we only need workstations to render a ground truth training dataset, saving the cost and hassle of building a physical one. Second, since photogrammetry-based face scans are much easier to acquire than OLAT scans, we are able to create a more diverse dataset with a higher diversity of ages, genders, and ethnicity. Finally, rendering a synthetic dataset also provides the ground truth for several other useful attributes. We show that this can enable new capabilities, such as controlling glares in eyeglasses in Section~\ref{sec:exp}. We will release our synthetic dataset upon publication to support other researchers in conducting portrait relighting research.

Although the virtual light stage provides a way to \YY{learn} the \YY{physically-accurate relighting behaviors} \YY{from synthetic data}, the realism of the deep network output is often limited when trained with only synthetic data. After all, the photogrammetry face scan comes with only basic reflectance information such as diffuse albedo and does not provide all the necessary information for realistic portrait relighting, such as surface~\cite{graham2013,nagano2015} and subsurface~\cite{henrik2001} reflectance properties. As a result, the generated synthetic training data exhibits a domain gap from their real counterparts. To close this gap, we propose a synthetic-to-real adaptation approach. We use in-the-wild portrait images to adapt the intermediate representations learned in our portrait relighting network for achieving photorealistic outputs. A specific challenge for the adaptation is how to improve photorealism while still maintaining the ability to relight an image coherently with the provided lighting condition. To this end, we propose self-supervised lighting consistency losses that promote the network to maintain consistent relighting during domain adaptation. We show that our approach leads to competitive portrait relighting performance to the state-of-the-art.

We further extend our relighting framework to video portrait relighting. Instead of applying relighting in a frame-by-frame manner, which often leads to flickering artifacts, we model the extension from image to video as a domain adaptation problem, following the same spirit of our synthetic-to-real approach. We use in-the-wild videos to adapt the intermediate representations learned in our portrait relighting network to improve temporal consistency. This results in a video portrait relighting network that is useful for various video applications.

In summary, our contributions include the following:
\begin{itemize}
    \item We propose a learning-based portrait relighting method that can achieve state-of-the-art results for challenging in-the-wild photos without requiring training data from a light stage.
    \item Our approach learns to approximate the physics for portrait relighting from a synthetic dataset, carefully constructed using a physically-based rendering engine. We further propose a novel synthetic-to-real adaption approach that leverages in-the-wild photos and a set of self-supervised lighting consistency losses to achieve photorealistic outputs.
    \item We extend our approach to portrait video relighting. We show that videos relit by our framework contain much less flickering and look temporally smoother than existing video relighting approaches.
\end{itemize}

The paper is organized as follows. We review related works in Section~\ref{sec:related_work}. Section~\ref{sec:dataset} presents our training data generation via a virtual light stage. Section~\ref{sec:method} gives details of our relighting network design and its adaptation from synthetic to real and from images to videos. Experiment results are included in Section~\ref{sec:exp}. Finally, Section~\ref{sec:conclusion} concludes the paper.

\section{Related work}\label{sec:related_work}

\subsection{Portrait Relighting}

Debevec \etal~\cite{debevec2000acquiring} show portrait relighting can be achieved by projecting the environment map to the reflectance field constructed by the OLAT images captured with a light stage setup. Wenger \etal~\cite{wenger2005performance} further extend the framework to allow moderate subject motions during the OLAT image acquisition. These OLAT-based approaches can produce high-quality relighting results for the subject whose OLAT capture is available but cannot be applied to unseen subjects. Despite the limitation, these methods are valuable for building the ground truth dataset for learning-based portrait relighting methods discussed below.

Various deep neural network architectures exist for portrait relighting based on directional lighting~\cite{nestmeyer2020faceRelighting}, and HDR environment map lighting~\cite{sun2019single,pandey2021total,mallikarjun2021photoapp,zhang2021neuralvideo}. 
SIPR-S~\cite{sun2019single} uses an encoder-decoder architecture. Total Relighting (TR)~\cite{pandey2021total} combines a set of UNets to generate the final output and achieves state-of-the-art results. Zhang \etal~\cite{zhang2021neuralvideo} propose a network architecture for video portrait relighting. \YY{NLT~\cite{zhang2021neurallight} learns a 6-DoF light transport function of UV locations, lighting directions, and viewing directions to enable novel view synthesis and relighting from OLAT images.} Our architecture is based on the TR framework, but we have made several improvements. First, we modify it to leverage supervision from our own custom synthetic data. We also add novel modules to bridge the synthetic-to-real domain gap and achieve temporally consistent video portrait relighting results.

Similar to ours, SIPR-W~\cite{wang2020single} uses captured face scans to create synthetic data to train portrait relighting networks. However, their synthetic humans have fixed looking. They do not pair them with different hair, clothes, and accessories. As a result, the diversity of the training data is limited, which throttles the relighting network performance. Moreover, they do not handle the synthetic-to-real domain gap and fall short in achieving photorealistic outputs. More recently, Sengupta \etal~\cite{sengupta2021light} propose a lightweight capture system to record facial appearance using only an LED monitor at the desk, circumventing the need for a light stage. However, their  monitor-based capture setup constrains their method to limited brightness controls and head motions, and near and frontal light sources. Other deep learning-based portrait relighting methods exist that do not require high-quality ground truth data. Still, they assume spherical harmonics (SH) lighting ~\cite{sengupta2018sfsnet,zhou2019deep,hou2021towards} that limit their method to low-frequency Lambertian surfaces. Our method adapts HDR image-based relighting representation that can handle arbitrary natural environments, including specular highlights and cast shadows.

\subsection{Learning with Synthetic Datasets}

Synthetic data provide valuable supervision for tasks where acquiring ground truth real data is challenging. They have been shown powerful for various tasks, including pose estimation~\cite{shotton2011real}, face analysis~\cite{wood2021fake}, pedestrian detection~\cite{fabbri2021motsynth}, segmentation~\cite{richter2016playing}, inverse rendering~\cite{li2020inverse,li2021openrooms}, 3D reconstruction~\cite{li2020through,wood20223d}, \YY{single-image full-body relighting~\cite{kanamori2019relighting,lagunas2021single} and scene relighting~\cite{philip2019multi,philip2021free}}. Our work is along this line. We show synthetic data are useful in producing state-of-the-art relighting results.

One drawback of using synthetic data to train a learning-based approach is the exposure to the synthetic-to-real domain gap problem. After all, synthetic data and real data follow two different distributions. To bridge the gap, many prior works develop methods that leverage domain knowledge in their applications. Representative approaches include synthetic-to-real for semantic segmentation~\cite{dundar2018domain,hoffman2018cycada}, 3D human pose estimation~\cite{doersch2019sim2real}, animal pose estimation~\cite{li2021synthetic}, GTA-to-real~\cite{richter2022enhancing}, SVBRDF estimation~\cite{vecchio2021surfacenet}, depth estimation~\cite{atapour2018real,zheng2018t2net}, vehicle re-Id \cite{lee2020strdan}, \YY{and multi-view portrait view synthesis and relighting~\cite{sun2021nelf}}. We follow the same spirit and devise a synthetic-to-real approach dedicated to the \YY{single-view} portrait relighting task. In a closely related work, 
Tajima~\etal~\cite{tajima2021relighting} propose a two-stage method for single-image full-body relighting with synthetic-to-real domain adaptation. In the first stage, they train a neural network for diffuse-only relighting using a spherical harmonic-based lighting representation. In the second stage, they train a network for enhancing non-diffuse reflection by learning residuals between real photos and images reconstructed by the previous-trained diffuse-only network. Our method differs from theirs in several aspects. First, we focus on portrait images that contain high-resolution facial details where errors are less forgiving. Second, our method takes environment maps as the lighting representation for image-based relighting, which is more expressive. Third but not least, we develop a novel self-supervised lighting consistency loss to achieve high-quality results.

\section{Synthetic Dataset Generation}
\label{sec:dataset}
\begin{figure*}[t]
\centering
\includegraphics[width=\linewidth]{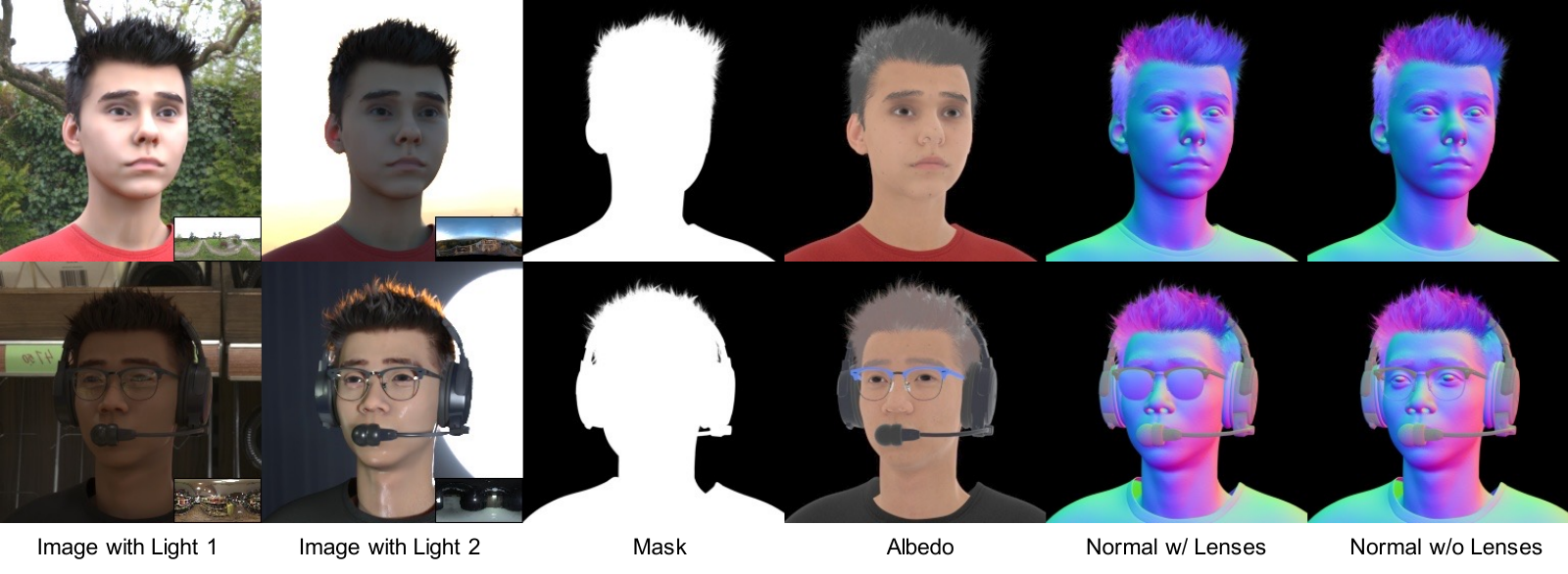}
    \vspace{-0.6cm}
  \caption{Examples of our dataset rendering. Our rendering consists of paired images under different lighting using HDR maps (as shown in the bottom right insets), along with the corresponding foreground mask, albedo, and normal maps. In addition, we render normal maps with lenses and without lenses to better model the reflections on glasses. Additional rendered attributes such as diffuse, specular, and shadow maps can be found in Appendix~\ref{sec_a:dataset}.
  }
\label{fig:rendering}
\end{figure*}

\begin{figure*}[t]
\centering
\includegraphics[width=\linewidth]{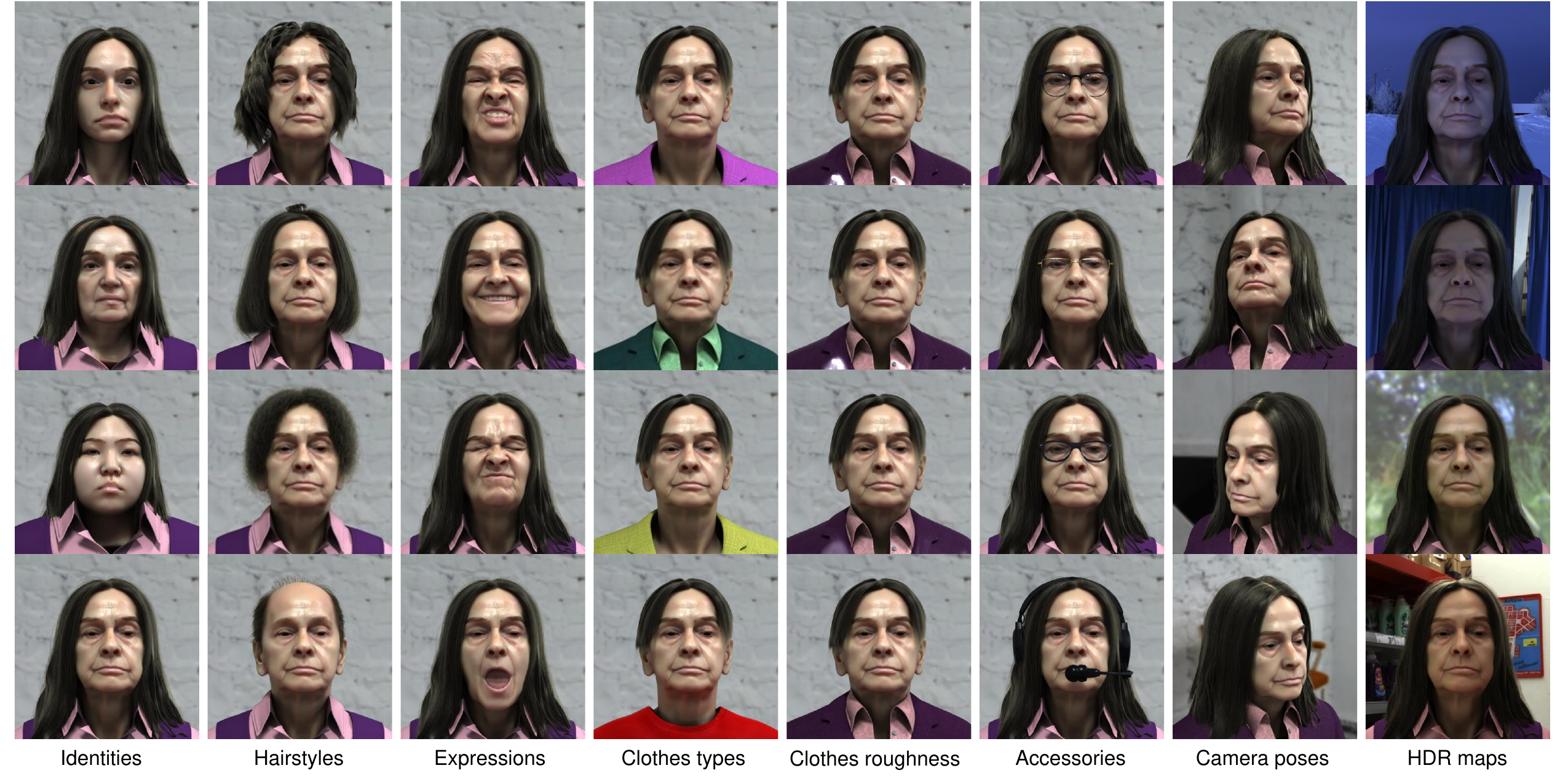}
    \vspace{-0.6cm}
  \caption{Examples of the variations in our dataset rendering. For each column, we change one variable which is specified at the bottom.}
\label{fig:variations}
\end{figure*}

For training a portrait relighting network, paired data are required for supervision. In the dataset, each sample consists of a pair of images of the same person at the same pose and expression but under two different lighting conditions. We generate the paired training dataset with a virtual light stage and a synthetic human generation pipeline described below.

Our synthetic human generation pipeline leverages pre-captured face scans that contain various identities in different facial expressions. Note that these face scans come with neither hair nor accessories. We thus randomly pair them with strand hair models of different styles and with various accessories and clothes. Finally, we put the synthetic humans in a virtual light stage where we illuminate them using HDR environment maps. The virtual light stage rendering is performed using a physically-based path-tracing renderer to simulate the light transport faithfully.

\begin{table*}
\begin{center}
\caption{Dataset comparisons for different portrait relighting methods. SIPR-S~\cite{sun2019single}, TR~\cite{pandey2021total} and NVPR~\cite{zhang2021neuralvideo} use a light stage to capture OLAT sequences and generate relit ground truth images. SIPR-W~\cite{wang2020single} and Ours acquire a large number of 3D face scans and render synthetic datasets by applying HDR maps as the environment lighting.
}
\vspace{-0.2cm}
\label{tab:baseline}

\begin{tabular}{cccccccccc}
\toprule
& Data acquisition & \multicolumn{2}{c}{Subjects} & \multicolumn{3}{c}{Add-ons per subject} & \multicolumn{2}{c}{HDR maps} \\
& & Quantity & Genders & Hairstyles & Clothes & Accessories & Real & Synthetic \\ 
\cmidrule(r){2-2} \cmidrule(r){3-4} \cmidrule(r){5-7} \cmidrule(r){8-9}
SIPR-S & light stage & 22  & 17M, 5F    & single   & single   & single   & 3094 & $\times$     \\ 
TR     & light stage & 70  &    -      & single   & multiple & multiple & 200  & $\times$     \\ 
NVPR   & light stage & 36  & 18M, 18F   & single   & single   & single   & 2810 & $\times$     \\ \cmidrule(r){1-9}
SIPR-W & simulation  & 108 &     -     &  none    & none     & none     & 391  & $\times$     \\ 
Ours   & simulation  & 515 & 247M, 265F & multiple & multiple & multiple & 1606 & \checkmark
\\  \bottomrule
\end{tabular}

\end{center}
\end{table*}

\subsection{3D Face Scans}
\label{sec:dataset:face}

We use high-quality 3D photogrammetry-based face scans with high-resolution texture maps for our synthetic humans. We note that these assets can be captured through a relatively low-cost system~\cite{Bee10} based on a passive camera array, which is much cheaper than a light stage. Moreover, there are already commercially available high-quality 3D face scans. The acquisition cost is lower than building a physical light stage. We acquired a commercial 3D face scan collection from the Triplegangers website\footnote{\url{https://triplegangers.com/}}. Each face scan is associated with a high-quality 4k diffuse texture and a high-resolution face geometry to represent mesoscopic facial details. Since the output mesh from a multi-view stereo process is unstructured, we fit our template model of common topology with 12325 vertices to the scan with non-rigid ICP and represent the remaining facial geometry details using a 4K displacement map. \YY{This common template registration is necessary to automate the synthetic data generation processes, such as fitting various hairstyles, applying common rigid transformations, and apply common UV-space material maps.} We use the Arnold \textit{aiStandardSurface}\footnote{https://docs.arnoldrenderer.com/display/A5AFMUG/Standard+Surface} shader to describe the face material, which includes path-traced subsurface scattering and specular highlights. We set the diffuse texture as the subsurface scattering color. We also apply different materials to different eye regions to render the specularity of eyes. Details of face materials can be found in Section~\ref{sec_a:dataset} of the Appendix.

The Triplegangers dataset consists of $248$ males and $267$ females of unique identities. Each identity has up to $20$ facial expression variations, leading to over 10k face scans in total. It also features a diverse range of ages and ethnicity, including ages ranging from 18 to 76-year-old and six ethnic groups. The detailed statistics can be found on the Triplegangers website.

\subsection{Hair}
\label{sec:dataset:hair}

Since our face scans do not contain hair, we use XGen~\cite{thompson2003xgen} in Maya\footnote{\url{https://www.autodesk.com/products/maya/}} to model our own custom hair strand templates. Each hairstyle is created by an experienced artist and is composed of several XGen descriptors for different hair portions, such as eyelashes, eyebrows, and beards. The Arnold aiStandardHair\footnote{\url{https://docs.arnoldrenderer.com/display/A5AFMUG/Standard+Hair}}, a physically-based hair shader, is applied to these XGen descriptors. In total, we have $24$ hairstyles, and each of them can be branched into many variations. To further increase the diversity, we uniformly sample the \textit{melanin} (within (0, 1)) and \textit{melanin redness} (within (0, 1)) to change hair colors. In our dataset, each hairstyle is fitted to various head scalps of 3D face scans.

\subsection{Clothing and Accessories}
\label{sec:dataset:clothes}

We acquire seven 3D garment models from the Turbosquid website\footnote{\url{https://www.turbosquid.com}} to add clothing variations to our synthetic humans. These seven garments have different styles, ranging from business shirts to T-Shirts. In addition, we randomize the color and the roughness of clothes (from 0.1 to 1.5) to cover a wider range of clothing.

We also acquire three different types of eyeglasses and one headphone from the Turbosquid website. These accessories are randomly added to our synthetic humans during rendering. These assets are added since they are most common for video conferencing, which is one of the major applications for single-image portrait relighting. In the future, we plan to add other accessories such as earring, necklace, scarf, and mask.

\subsection{Rendering} \label{sec:dataset:render}

We use Arnold~\cite{georgiev2018arnold} in Maya, a physically-based path tracer, to render our training dataset. The renderer simulates the light transport based on physics and can create realistic outputs. 

The training dataset consists of paired samples. Each sample consists of a synthetic human with the same pose and expression rendered under two different lighting. We create a synthetic human by sampling a face scan, a hairstyle, a garment, a pair of eyeglasses (at a probability of $0.2$), and a headset (at a probability of $0.05$). The synthetic human is positioned near the world center with some randomness in the precise location. We sample the camera pose from a sphere centered at the world center with a fixed radius. The viewing angle is restricted to within a range of $(-45^{\circ},45^{\circ})$ horizontally and $(-22.5^{\circ},22.5^{\circ})$ vertically in front of the faces. The camera pose variations result in variations in head poses in the rendered images of the synthetic humans.

Once a synthetic human is created, and a camera pose is sampled, we use two different environment maps to illuminate it to create the paired sample. We use latitude-longitude format HDR images as our lighting. We acquire $1536$ HDR images for training and $70$ images for evaluation from a variety of sources, including Poly Haven\footnote{\url{https://polyhaven.com}}. The HDR images are captured from various environments, including indoors and outdoors. To add more lighting variations, we apply random rotations and horizontal flips to the HDR images.

In addition to color images, we also render other attributes such as albedo and normal maps, which can be easily computed using our synthetic data generation pipeline. Moreover, we render two versions of surface normals, one with lenses in eyeglasses and one without. The double normal map treatment helps us synthesize glares in the eyeglasses. Examples of rendered results can be found in Figure~\ref{fig:rendering}. We demonstrate variations of hair, identities, expressions, clothes, and accessories available to a face scan in our dataset in Figure~\ref{fig:variations}. Overall, we rendered about 300k paired samples, where each sample contains a pair of RGB images and the attributes mentioned above. Each image has a resolution of $512 \times 512$ and is stored in the HDR format.
\section{Method} \label{sec:method}

\begin{figure}[t]
\centering
\includegraphics[width=\linewidth]{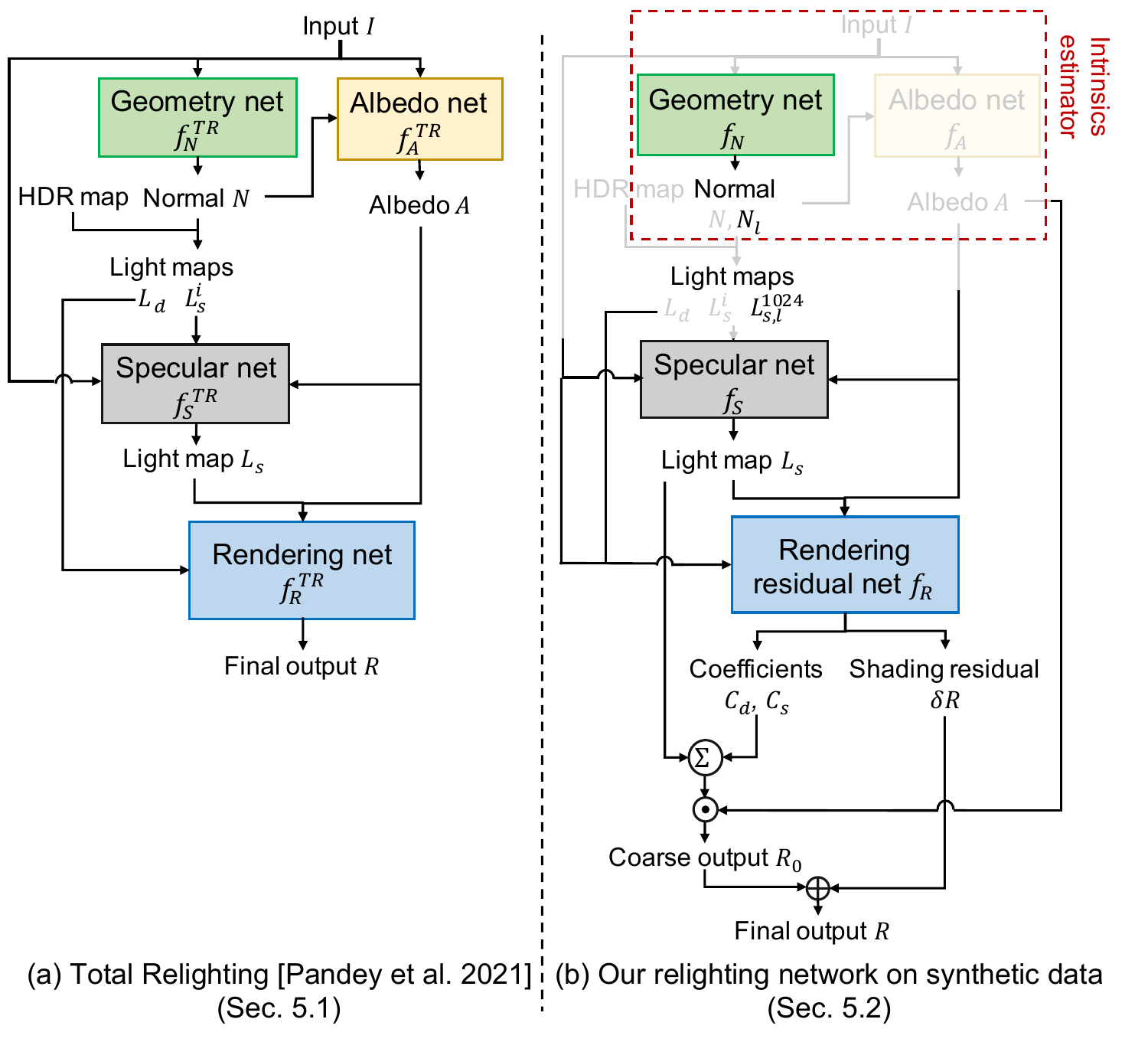}
    \vspace{-0.5cm}
  \caption{Our portrait relighting method when trained on the synthetic dataset. (a) The baseline Total Relighting network~\cite{pandey2021total}. (b) Compared with (a), we make two modifications. First, we predict an additional normal map $\lens{N}$ which contains glasses lenses, so it can be used to create a new highly specular light map ${L}_{s,l}$ for glares on glasses. Second, instead of using $f_R$ as a black box to generate the final output, we first obtain a coarse output $R_0$ using the estimated albedo and light maps. We then only estimate the fine details $\delta R$ to be added to this coarse output to obtain the final output. Note that the components similar to (a) are grayed out.}
  \vspace{-0.3cm}
\label{fig:method_base}
\end{figure}

\begin{figure}[t]
\centering
\includegraphics[width=\linewidth]{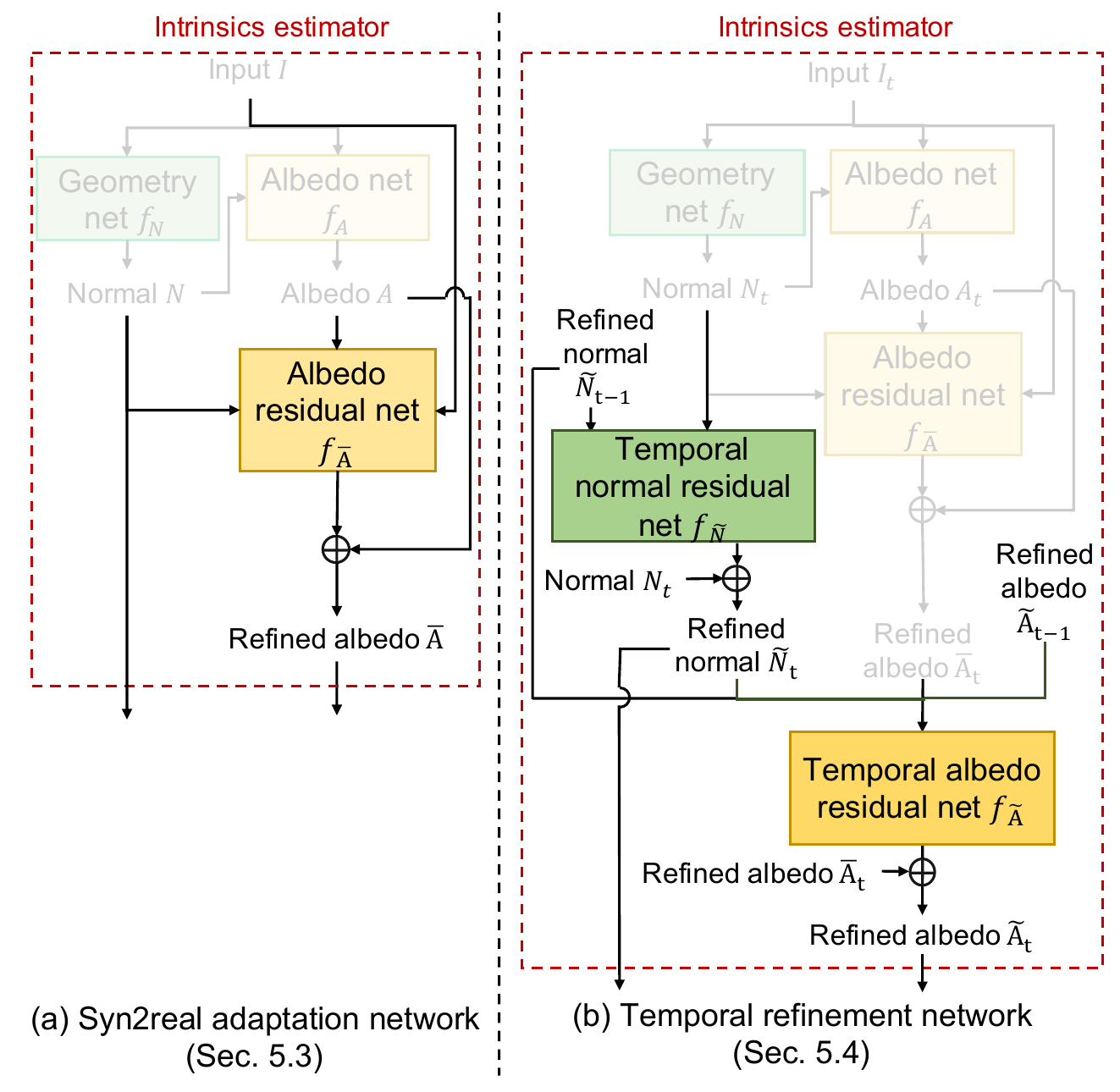}
\vspace{-0.5cm}
  \caption{Our syn2real adaptation and temporal refinement networks. Note that we only modify the intrinsics estimator block in Figure~\ref{fig:method_base}. (a) To bridge the domain gap between synthetic and real domains, we observe that the gap mainly comes from the estimated albedo map. We thus train a domain adaptation network to refine the albedo map by predicting an albedo residual, using only \emph{unpaired} real data. (b) Finally, to increase temporal stability when relighting videos, we can (optionally) train two more residual networks to refine the normal and albedo maps using real videos. For each subfigure, the part similar to the previous subfigure is grayed out.}
  \vspace{-0.3cm}
\label{fig:method_adapt}
\end{figure}

The full framework of our portrait relighting method is shown in Figures~\ref{fig:method_base} and~\ref{fig:method_adapt}. 
We first obtain the foreground image $I$ using an off-the-shelf matting network~\cite{MODNet} given an input portrait image. This image and an environment map are inputs to our relighting network to predict a relit image $R$.

Our relighting network is based on the TR framework~\cite{pandey2021total}. However, instead of adopting it directly, we made several modifications to adapt it to our scenario. Below we first briefly review the TR framework (Section~\ref{sec:method:pre}). We then explain how we modify it to train on our synthetic dataset (Section~\ref{sec:method:syn}). Next, we finetune the model on real images to bridge the synthetic-to-real domain gap (Section~\ref{sec:method:adapt}). Finally, we show how we can further finetune the model on real videos to increase temporal stability when relighting portrait videos (Section~\ref{sec:method:video}).

\begin{table*}[!bt]
\begin{center}
\caption{Notations used in this paper.}
\vspace{-0.2cm}
\label{tab:notation}

\begin{tabular}{cccccc}
\toprule
\multicolumn{2}{c}{Baseline} & \multicolumn{2}{c}{Syn2real} & \multicolumn{2}{c}{Temporal} \\
\cmidrule(r){1-2} \cmidrule(r){3-4} \cmidrule(r){5-6}
$I$ & Input foreground image \\
$N$ & Normal & $N_l$ & Normal with lenses & $\dbar{N}_t$ & Temporally-refined normal \\
    &        &       &                    & $\delta{\dbar{N}}_t$ &  Temporally-refined normal residual \\
$A$ & Albedo & $\sbar{A}$ & Refined albedo & $\dbar{A}_t$ & Temporally-refined albedo \\
    &        & $\delta{\sbar{A}}$ & Refined albedo residual & $\delta{\dbar{A}}_t$ & Temporally-refined albedo residual \\
$L_d$ & Diffuse light map \\
$L_s$ & Specular light map & $L_{s,l}$ & Specular light map w/ lenses \\
$W_s$ & Weight for specular light map & $W_{s,l}$ & Weight for specular light map w/ lenses \\
$R$ & Relit output image & $\sbar{R}$ & Refined relit output & $\dbar{R}_t$ & Temporally-refined relit output \\
$R_0$ &  Initial Rendering  &            &                       &           &                    \\
$\delta{R}$ &  Relit image residual  &            &                       &           &                    \\
\cmidrule(r){1-6}

$f_N$ & Normal net &&& $f_{\dbar{N}}$ & Temporally-refining normal net \\
$f_A$ & Albedo net & $f_{\sbar{A}}$ & Albedo refining net & $f_{\dbar{A}}$ & Temporally-refining albedo net \\
$f_S$ & Specular net &&&& \\
$f_R$ & Render net &&&& \\

\bottomrule

\end{tabular}
\end{center}
\vspace{-0.2cm}
\end{table*}

\subsection{Total Relighting~\cite{pandey2021total}}
\label{sec:method:pre}

At a high level, the TR framework works as follows: Given an image, it first predicts the normal and albedo maps. It then predicts the diffuse and specular light maps with respect to the target environment map illumination using the predicted normals. Finally, these components are concatenated as input to another network to produce the final relit image.
The framework consists of $4$ networks: a geometry network $f_N^{TR}$, an albedo network $f_A^{TR}$, a specular network $f_S^{TR}$, and a rendering network $f_R^{TR}$. These networks are all implemented using a UNet architecture.

The geometry network $f_N^{TR}$ first takes the foreground image $I$ as input, and generates a surface normal map $N$,
\begin{equation}
N = f_N^{TR}(I).
\end{equation}
Next, the foreground image $I$ is concatenated with the surface normal map $N$ as input and sent to the albedo network $f_A^{TR}$. The output is an albedo map $A$,
\begin{equation}
A = f_A^{TR}(I, N).
\end{equation}

For environment maps, a diffuse irradiance map and a set of prefiltered environment maps according to specular Phong exponents ($n=1,16,32,64$) are precomputed~\cite{miller1984illumination}. The prefiltering of environment maps makes the integration computation offline and thus enables the real-time rendering of diffuse and specular materials. Here the \textit{light maps} $L_d, L_s^1, L_s^{16}, L_s^{32}, L_s^{64}$ can be computed by simply indexing the prefiltered maps using the normal or reflection directions depending on predicted surface normals $N$. The diffuse and specular light maps are more efficient representations of lighting than the original environment map as they now embed the diffuse and specular components of illumination in the pixel space. 

The specular network $f_S^{TR}$ takes foreground image $I$, albedo map ${A}$, and four specular light maps $L_s^1, L_s^{16}, L_s^{32}, L_s^{64}$ as input, and predicts the four-channel per-pixel weight map $W_s^1, W_s^{16}, W_s^{32}, W_s^{64}$, 
\begin{equation} \label{eqn:spec_net}
W_s^1, W_s^{16}, W_s^{32}, W_s^{64} = f_S^{TR}(I, A, L_s^1, L_s^{16}, L_s^{32}, L_s^{64}).
\end{equation}
The weighted specular light map $L_s$ can be computed by $L_s=\sum_{i}W_s^i\odot L_s^i$, where $i=1, 16, 32, 64$ and $\odot$ indicates pixel-wise multiplication.

Finally, the rendering network $f_R^{TR}$ takes albedo map $A$, diffuse light map $L_d$, and specular light map $L_s$ as input and predicts the relit image $R$,
\begin{equation}
R = f_R^{TR}(A, L_d, L_s).
\end{equation}

Pandey \etal~\shortcite{pandey2021total} train the networks using $L_1$ losses, perceptual losses, and GAN losses on the normal, albedo, and specular maps, as well as the final output. Please refer to their paper for more details.

\subsection{Learning with Synthetic Data}
\label{sec:method:syn}

We make two major modifications to the TR framework when training it on our dataset. The first one is that we improve the rendering network through \YY{a decomposition scheme. Unlike previous work that employs the rendering decomposition (e.g., ~\cite{zhang2021neurallight, nagano2019deep}, we decompose the output to two \textit{learnable} components: coarse rendering by predicting a set of parameters for a fixed rendering function and fine-scale residuals.} The second modification is that we predict the normal for the lenses of the glasses prediction.

\begin{figure}[t]
\centering
\includegraphics[width=\linewidth]{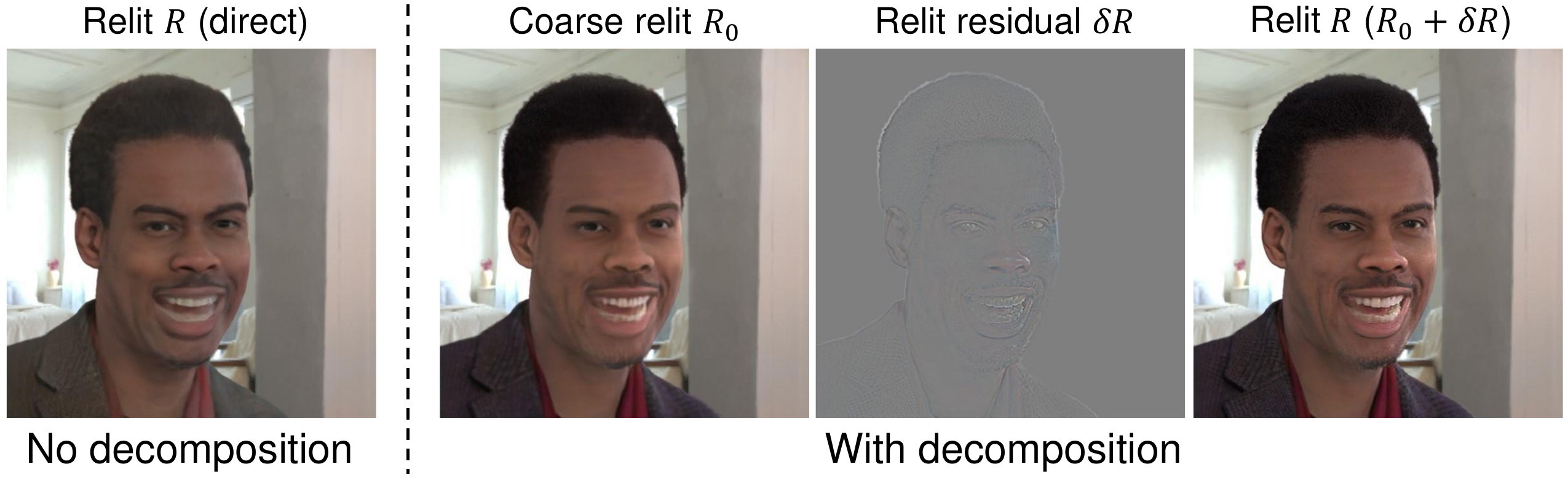}
    \vspace{-0.6cm}
  \caption{An example showing using a blackbox for final rendering vs.\ using an initial rendering plus adding fine details (e.g., teeth) as in our framework.}
  \vspace{-0.3cm}
\label{fig:decomposition}
\end{figure}

\paragraph{Improved rendering network.}
Different from the TR framework, which treats the rendering network $f_R$ as a black box to get the final relit image, we first compute an initial rendering from the predicted albedo, diffuse, and specular light maps. We then refine the result by only estimating the details which cannot be modeled by this coarse rendering. Specifically, the final rendering is decomposed into two steps. We first predict coefficients to linearly combine the diffuse and specular light maps and then multiply the combined map by the albedo map to obtain an initial result,
\begin{equation}
\label{eqn:init_render}
 R_0=A\odot(C_d \cdot L_d + C_s \cdot L_s),
\end{equation} 
where $\odot$ stands for element-wise multiplication, and $C_d$, $C_s$ are the predicted 3-channel coefficients. We then estimate a residual map $\delta R$ to account for the details not modeled by this initial rendering. The final relit image can then be computed by ${R}={R}_0+\delta{R}$. An example of our decomposition of initial coarse rendering $R_0$, residual image $\delta{R}$, and the final relit image $R$ is shown in Figure~\ref{fig:decomposition}. The facial details cannot be retained well if the final rendering is predicted by a blackbox without decomposition.

In our framework, the coefficients and the residual map are predicted by $f_{R}$, which takes the foreground image $I$, the albedo map ${A}$, the diffuse light map ${L}_d$, and the specular light map ${L}_s$ as input,
\begin{equation}
{C}_d, {C}_s, \delta{R} = f_{R}(I, {A}, {L}_d, {L}_s).
\end{equation}

Similar to the TR framework, we employ losses on the normal, albedo, and specular maps, as well as the final output. In addition, to avoid the residual map $\delta{R}$ becoming too large and completely dominating the coarse rendering output ${R}_0$, we add an $L_1$ regularization on $\delta{R}$, i.e., $\mathcal{L}_{\delta R} = ||\delta{R}||_1$.

Example results by applying this modification are shown in Figure~\ref{fig:ablation}. Although this does not lead to large-quality improvement when trained on the synthetic dataset, it has significant effects when applying the synthetic-to-real adaptation to real data, as introduced in Section~\ref{sec:method:adapt}.

\paragraph{Normal prediction for glasses.} Since our synthetic dataset contains normal maps both with lenses and without lenses for glasses, we train our network to predict both of them. In particular, our geometry network $f_{N}$ generates two normals,
\begin{equation}
    N, \lens{N} = f_{N}(I),
\end{equation}
where $N$ is the normal map without lenses and $\lens{N}$ is the normal map with lenses. \YY{If input has no eyeglasses, we predict the same normal maps for $N$ and $\lens{N}$.} Then, when generating the specular light maps, in addition to the four original specular exponents ($L_s^1, L_s^{16}, L_s^{32}, L_s^{64}$) using $N$, we generate an additional highly specular one ${L}_{s,l}^{1024}$ using $\lens{N}$ to account for the glares. Our specular network $f_{{S}}$ then predicts an additional weight ${W}_{s,l}^{1024}$ compared to Equation.~\ref{eqn:spec_net},
\begin{equation}
W_s^1, W_s^{16}, W_s^{32}, W_s^{64}, {W}_{s,l}^{1024} = f_{{S}}(I, A, L_s^1, L_s^{16}, L_s^{32}, L_s^{64}, {L}_{s,l}^{1024}).
\end{equation}
The weighted specular light map then becomes $L_s=\sum_{i}W_s^i\odot L_s^i + {W}_{s,l}^{1024}\odot {L}_{s,l}^{1024}$ where $i=1, 16, 32, 64$. This allows us to enable the eyeglasses glare controlling feature, as demonstrated in Figure~\ref{fig:teaser}, Figure~\ref{fig:glasses_glares}, and the supplementary video.

\subsection{Synthetic-to-Real Adaptation on Real Data}
\label{sec:method:adapt}

\begin{figure}[t]
\centering
\includegraphics[width=\linewidth]{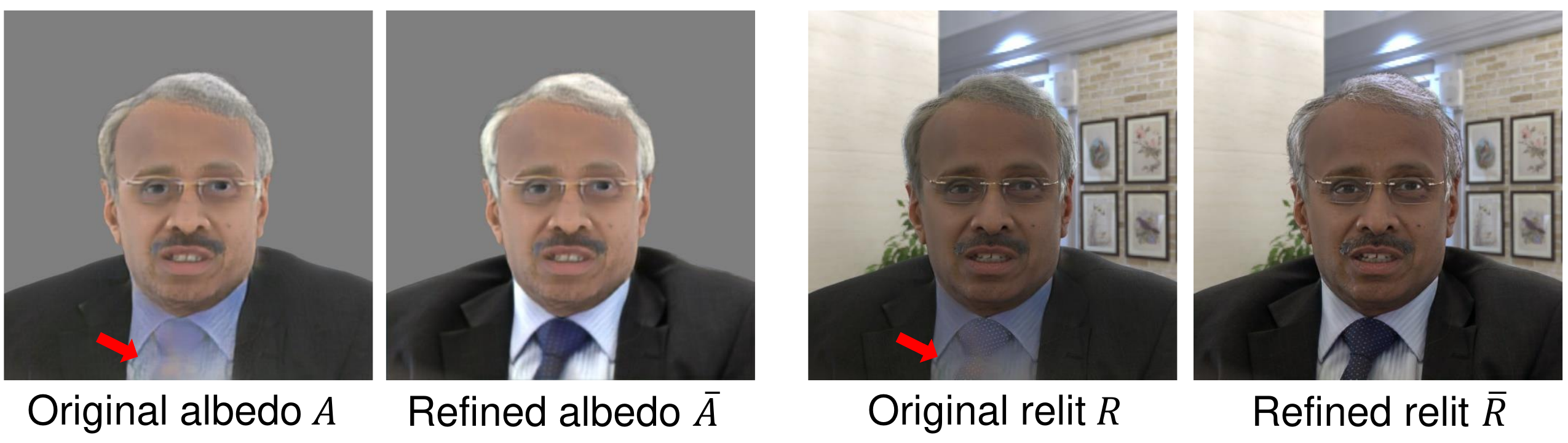}
    \vspace{-0.6cm}
  \caption{An example showing the initial vs.\ refined albedo and their corresponding relit outputs. Note how the refined albedo and relit output generate much better clothing.
  }
  \vspace{-0.3cm}
\label{fig:residuals}
\end{figure}

Although the model trained in Section~\ref{sec:method:syn} using our synthetic dataset can generate reasonable relit results, the outputs often lack photorealism when tested on real images. This is because there is a domain gap between rendered and real-world portrait photos. To bridge this gap, we propose a novel synthetic-to-real (syn2real) mechanism for portrait relighting.

In particular, instead of naively applying an existing syn2real network on the final relit image, we observed that the domain gap between synthetic and real data mainly comes from the diversity of the subject albedo, which strongly determines the final appearance in the relit image. People have more diverse hairstyles, face colors, minor facial marks, wearings, and accessories in the real world, which are missing in the synthetic data. This causes the albedo network trained with only synthetic data to fail when modeling real-world appearances.
\YY{We tried a few variants of the adaptations including the adaptation of surface normals, but found empirically that the proposed albedo adaptation framework works better with our synthetic dataset.}

To handle the domain gap between the albedo maps, we add an additional albedo residual network $f_{\sbar{A}}$ to refine the initially estimated albedo. The network $f_{\sbar{A}}$ takes the portrait image $I$, the normal map ${N}$, and the estimated albedo ${A}$ as input and outputs an albedo residual $\delta{\sbar{A}}$,
\begin{equation}
\delta{\sbar{A}} = f_{\sbar{A}}(I, {N}, {A}).
\end{equation}
The refined albedo, $\sbar{A}={A}+\delta{\sbar{A}}$, then replaces the original albedo ${A}$ as the input to the rest of the model ($f_S$ and $f_{\delta R}$) to predict a new relit output $\sbar{R}$. As shown in Figure~\ref{fig:residuals}, the refined albedo after adaptation can recover input details (cloth color in this case) that are not seen in synthetic data.

When finetuning on real data, we only train $f_{\sbar{A}}$ while keeping the rest of the model fixed. The loss functions include a similarity loss, an identity loss, a GAN loss, and two lighting consistency losses. For all losses, the superscript refers to the loss type, while the subscript refers to the output where the loss is computed on.

\begin{figure}[t]
\centering
\includegraphics[width=\linewidth]{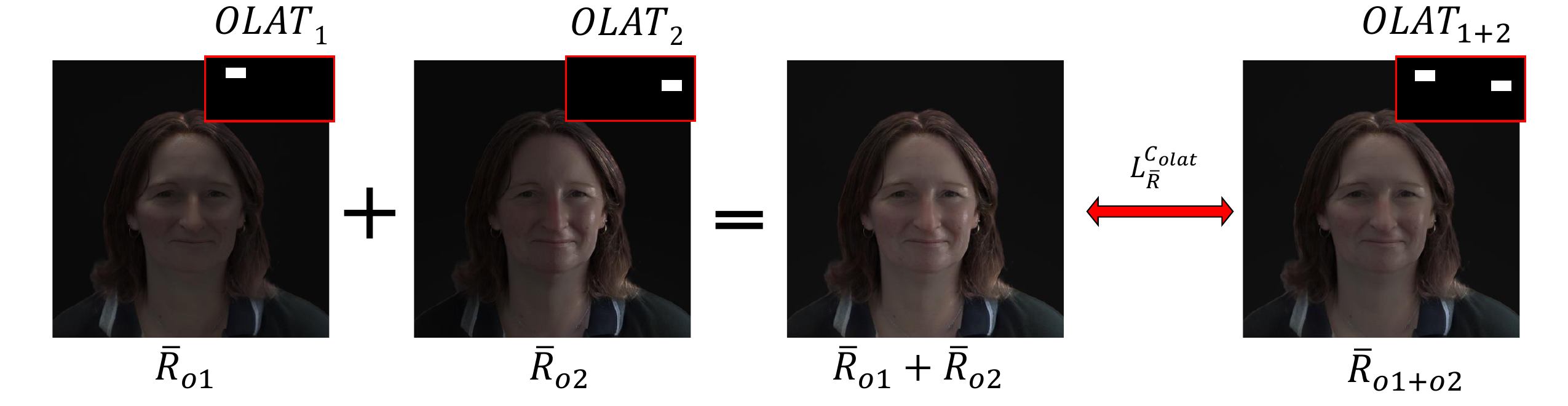}
    \vspace{-0.6cm}
  \caption{
  OLAT consistency loss. We relight the input using two \YY{randomly picked} OLAT maps and compute the difference between the sum of them and the result when we relight using the sum of the two OLAT maps.}
  \vspace{-0.3cm}
\label{fig:consistency_olat}
\end{figure}

\begin{figure}[t]
\includegraphics[width=\linewidth]{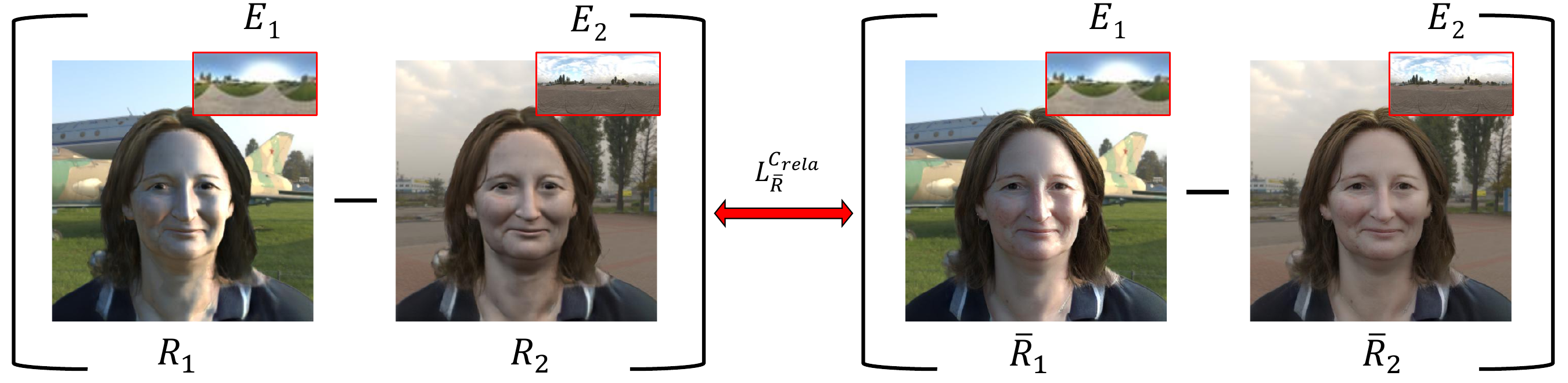}
    \vspace{-0.6cm}
  \caption{
  Relative consistency loss. We relight the input using two \YY{randomly picked} environment maps and compute the difference between the two relit outputs. This is done for both before and after the syn2real adaptation, and we compute the loss between the two differences.}
  \vspace{-0.3cm}
\label{fig:consistency_diff}
\end{figure}
\paragraph{Lighting consistency loss $\mathcal{L}_{\sbar{R}}^{C_{olat}}$, $\mathcal{L}_{\sbar{R}}^{C_{rela}}$.}
To keep the relighting effect consistent before and after the syn2real adaptation, we apply the following consistency losses between the relit image $\sbar{R}$ and the input foreground image $I$ \YY{as regularization terms}.

First, we use the \textit{OLAT lighting consistency loss}. This loss assumes that the illuminations and relit outputs have the distributive property, i.e., summing two environment maps and relighting with the sum leads to the same result as summing the two relit outputs, as shown in Figure~\ref{fig:consistency_olat}. In particular, we first generate a set of OLAT environment maps, where only one pixel is lightened in each of them to simulate the light stage scenario. Examples of OLAT environment maps can be found in Appendix~\ref{sec_a:olat}. During each training iteration, we randomly sample two of them ($OLAT_1$ and $OLAT_2$) to simulate two different light source locations and get the combined environment map $OLAT_{1+2}$, which simulates two light sources turned on at the same time. Let the output images relit by maps $OLAT_1$, $OLAT_2$, and $OLAT_{1+2}$ be $\sbar{R}_{o1}$, $\sbar{R}_{o2}$, and $\sbar{R}_{o1+o2}$, respectively. The loss is then
\begin{equation}
\label{eqn:olat_loss}
\mathcal{L}_{\sbar{R}}^{C_{olat}} = ||(\sbar{R}_{o1} + \sbar{R}_{o2}) - \sbar{R}_{o1+o2}||_1 .
\end{equation}
\YY{Using two OLAT images, instead of two environment maps, ensures that the relighting can react to high-frequency point lighting conditions such as the OLAT lighting as shown in the accompanying video.}

\begin{figure*}[t]
\centering
\includegraphics[width=\linewidth,trim=0 7pt 0 0, clip]{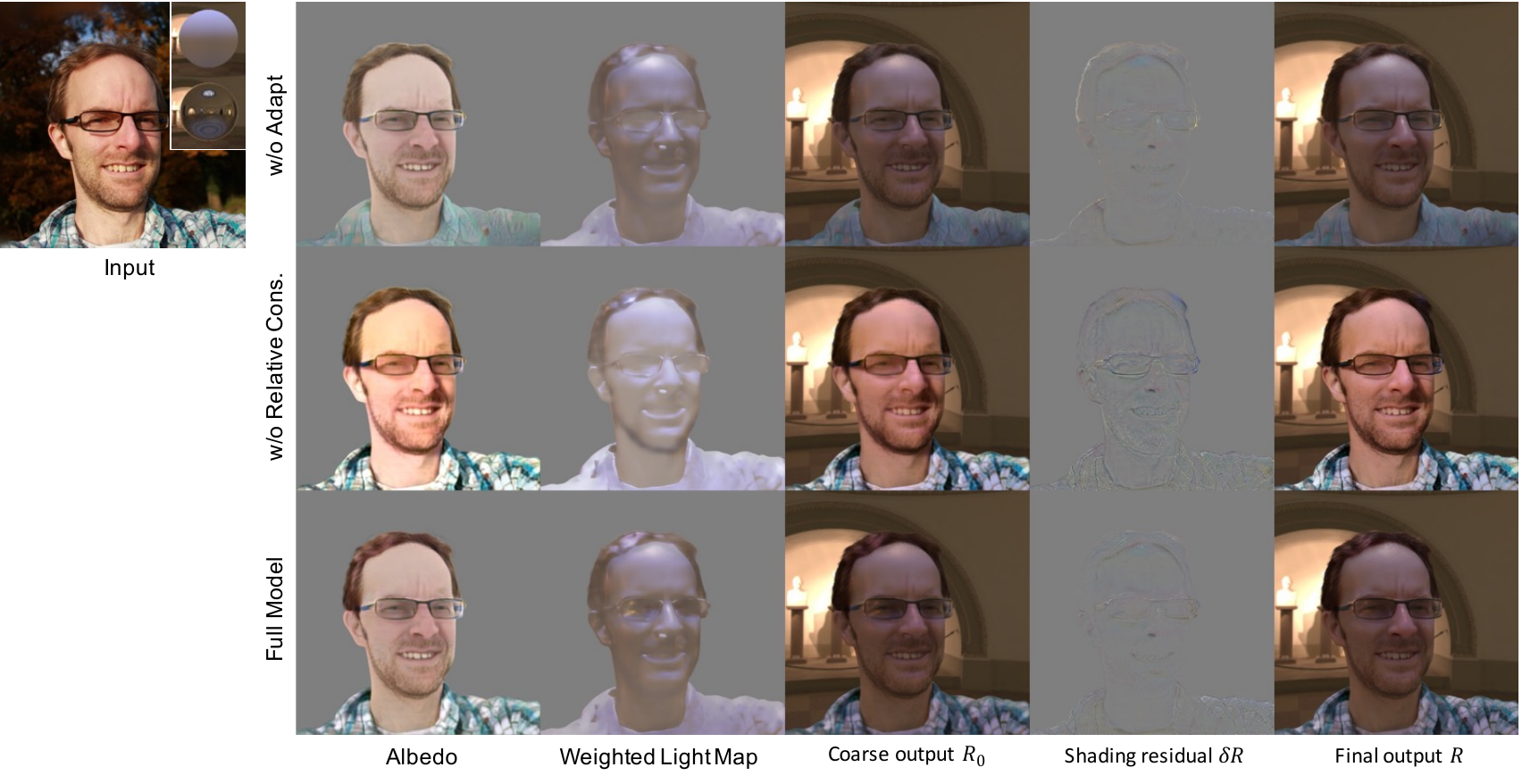}
    \vspace{-0.3cm}
  \caption{\YY{Comparisons on the effectiveness of the albedo adaptation and relative lighting consistency loss for the output layers. Without adaptation, the model cannot keep facial details such as hair colors and beards, and most notably the textures on the clothing. Without the relative lighting consistency loss, the input image lighting is baked into the predicted albedo and leads to inaccurate relighting results. Our full model can recover correct albedo colors and details better, since the proposed \textit{relative lighting consistency loss} aims to improve the photorealism in isolation under arbitrary target lightings.}  }
  \vspace{-0.3cm}
\label{fig:ablation_albedo}
\end{figure*}

Second, we use the \textit{relative lighting consistency loss}. This loss assumes that the adaptation should only change the photorealism \YY{and acts as a form of regularization to} preserve the \YY{relighting behaviors} that our network has learned from a physically-based renderer. As we do not have ground truth lighting, we enforce this idea through a pair of relighting outputs. \YY{Let $E_i$ and $E_j$ be an arbitrary pair of two environment maps randomly picked every time the loss is evaluated. We denote the relit results using $E_i$ and $E_j$ as ${R}_i$, ${R}_j$ before adaptation and $\sbar{R}_i$, $\sbar{R}_j$ after adaptation.} We then apply an $L_1$ loss between the two differences,
\begin{equation}
\mathcal{L}_{\sbar{R}}^{C_{rela}} = || ({R}_i-{R}_j) - (\sbar{R}_i-\sbar{R}_j) ||_1 .
\end{equation}
This way, the network has a stronger incentive to preserve the lighting produced before the syn2real adaptation, as the deviations from the original outputs will be discouraged. \YY{Mathematically, $||(R_i-R_j)-(\bar{R}_i-\bar{R}_j)||_1 = ||(R_i-\bar{R}_i)-(R_j-\bar{R}_j)||_1 = ||d_i-d_j||_1$. This encourages the improvements $d$ to be invariant to lighting conditions. It means that our loss will keep the overall lighting consistent before and after the adaptation but only change the photorealism.} We visualize the effect of the relative lighting consistency loss in Figure~\ref{fig:consistency_diff} \YY{and Figure~\ref{fig:ablation_albedo}}.
\YY{We note that the relative consistency loss plays an important role in our adaptation framework. Figure~\ref{fig:ablation_albedo} compares three variants: without the adaptation (\textit{w/o adaptation}), without relative lighting consistency loss (\textit{w/o relative consistency loss}), and our full model. As can be seen, \textit{w/o adaptation} (first row) leads to washed out albedo details and textures (e.g., on clothing). With adaptation, but without the relative lighting consistency (second row), the method fails to recover the correct albedo. For example, the cast sunlight in the input is baked into the predicted albedo and the model is unable to disentangle the lighting and the albedo in the input. With our full model (third row), both the details and correct albedo colors can be recovered after adaptation.}

\paragraph{Similarity loss $\mathcal{L}_{\sbar{A}}^{sim}$.}
To prevent the refined albedo from deviating too much from the original albedo, we extract VGG features from the original and refined albedo and put an $L_1$ loss between them.

\paragraph{Identity loss $\mathcal{L}_{\sbar{R}}^{Id}$.} 
To ensure the identity does not get changed after relighting, we apply a face recognition network~\cite{parkhi2015deep} on the input $I$ and relit images $\sbar{R}$, extract high-level features from them, and compute the $L_1$ distance between these two features.

\paragraph{GAN loss $\mathcal{L}_{\sbar{R}}^{G}$.} 
To make the relit images look realistic, we also put a GAN loss on the relit images $\sbar{R}$.

The overall loss is then
\begin{align}
    \mathcal{L}^{adap} = &\lambda^{C_{olat}}\mathcal{L}_{\sbar{R}}^{C_{olat}} + \lambda^{C_{rela}}\mathcal{L}_{\sbar{R}}^{C_{rela}} \nonumber\\
    + &\lambda^{sim}\mathcal{L}_{\sbar{A}}^{sim} + \lambda^{Id}\mathcal{L}_{\sbar{R}}^{Id} + \lambda^{G}\mathcal{L}_{\sbar{R}}^{G}
\end{align}
where $\lambda$'s are the weights and are set to $10$, $10$, $10$, $1$, $1$ respectively in our framework.

Example results before and after the domain adaptation can be found in Figure~\ref{fig:ablation}. We also show comparisons with directly applying syn2real on the output relit image as well as without adopting our improvement in Section~\ref{sec:method:syn}. \YY{The losses are the same as the full model, while the similarity loss is computed on the relit image instead of the albedo}. We found that both alternative approaches give the network too much ``freedom'' to change the relit output. As a result, the correct lighting information is not preserved after adaptation. In particular, when directly applying syn2real to the output, it creates a ``shortcut'' that the network just mimics the real image since the real image is both the input to the network and the target. On the other hand, when we do not adopt the modification in Section~\ref{sec:method:syn} and directly estimate the output using a black box $f_R$, the network tends to use the albedo map in an unpredictable way to generate the relit image. In our approach, the generation is fully controllable by the albedo map since a linear combination is used. 

\subsection{Relighting Portrait Videos} \label{sec:method:video}

In many applications, portrait video relighting is required. A straightforward way to achieve this capability is to simply apply the relighting model in a frame-by-frame manner. However, this usually results in flickering artifacts in the relit video. In particular, we observe that the flickering is due to the intermediate outputs in our framework changing too frequently between frames. These intermediate outputs include the estimated normal and albedo maps. To resolve the issue, we learn two temporal residual networks $f_{\dbar{N}}$ and $f_{\dbar{A}}$ to make the normal and albedo maps temporally consistent, respectively. Note that these temporal networks are only required for video portrait relighting and are learned while keeping the other parts of the framework fixed.

The temporal residual networks take outputs from the past frame and the predictions from the image-only relighting part of the framework at the current frame as inputs. In particular, $f_{\dbar{N}}$ takes the original normal prediction ${N}_t$ at time $t$ and the refined normal from the previous frame $\dbar{N}_{t-1}$ at time $t-1$ as input for predicting a temporal normal residual $\delta{\dbar{N}}_t$. This residual is added to the original normal prediction ${N}_t$ to obtain the refined normal ${\dbar{N}}_{t}$.
\begin{equation}
\delta{\dbar{N}}_t = f_{\dbar{N}}({\dbar{N}}_{t-1}, N_t),  \quad 
{\dbar{N}}_t = {N}_t + \delta{\dbar{N}}_t .
\end{equation}
Similarly, $f_{\dbar{A}}$ takes temporally refined normal ${\dbar{N}}_{t-1}$, ${\dbar{N}}_t$, temporally refined albedo ${\dbar{A}}_{t-1}$, and the albedo $\sbar{A}_t$ (after syn2real adaptation) as input, and predicts a temporal albedo residual $\delta{\dbar{A}}_t$. This residual is added to the syn2real albedo $\sbar{A}_t$ to obtain the temporally refined albedo ${\dbar{A}}_{t}$.

\begin{equation}
\delta{\dbar{A}}_t = f_{\dbar{A}}({\dbar{N}}_{t-1}, {\dbar{N}}_{t}, {\dbar{A}}_{t-1}, \sbar{A}_t),  \quad 
{\dbar{A}}_t = \sbar{A}_t + \delta{\dbar{A}}_t .
\end{equation}

After we obtain the refined normal and albedo maps, they are fed into the rest of the framework to obtain the final relit output $\dbar{R}_t$. To train the two networks, we again adopted the similarity loss and identity loss used in Section~\ref{sec:method:adapt}. To ensure temporal consistency, we also add a warping loss between neighboring frames. 

\paragraph{Similarity loss $\mathcal{L}_{\dbar{N}}^{sim}$, $\mathcal{L}_{\dbar{A}}^{sim}$.} Similar to our syn2real treatment in Section~\ref{sec:method:adapt}, we add a perceptual loss between the original (${N}_t$ and $\sbar{A}_t$) and temporally-refined attributes (${\dbar{N}}_t$ and ${\dbar{A}}_t$) to prevent the refined attributes from deviating too much from the original ones.

\paragraph{Identity loss $\mathcal{L}_{\dbar{R}}^{Id}$.} To ensure the identity does not get changed after refinement, we extract identity features from input $I_t$ and refined output $\dbar{R}_t$ using a face recognition network~\cite{parkhi2015deep} and put an $L_1$ loss between them.

\paragraph{Warping loss $\mathcal{L}_{\dbar{R}}^{W}$.}
We first compute the optical flow~\cite{flownet2-pytorch} on neighboring input frames $I_{t-1}$ and $I_t$. We then apply the flow to warp the previous relit frame ${\dbar{R}}_{t-1}$ and compute its difference to the current relit frame ${\dbar{R}}_{t}$,
\begin{equation}
  \mathcal{L}_{\dbar{R}}^{W} = \| \mathcal{W} ({\dbar{R}}_{t-1}) - {\dbar{R}}_{t} \|_1,
\end{equation}
where $\mathcal{W}$ is the warping function based on the estimated optical flow. The final loss is then
\begin{equation}
\mathcal{L}^{tem} = \lambda^{sim}\big(\mathcal{L}_{\dbar{N}}^{sim} +
\mathcal{L}_{\dbar{A}}^{sim}\big) + \lambda^{Id}\mathcal{L}_{\dbar{R}}^{Id} + \lambda^{W}\mathcal{L}_{\dbar{R}}^{W},
\end{equation}
where $\lambda$'s are loss weights and are set to $0.2, 0.1, 5$ in our framework.
In Figure~\ref{fig:comp_temporal}, we show that our temporal refinement can improve temporal consistency significantly.

\subsection{Datasets and Implementation Details}
For supervised learning of our portrait relighting network (Section~\ref{sec:method:syn}), we use our synthetic dataset created from Section~\ref{sec:dataset} for training. For syn2real finetuning on real data (Section~\ref{sec:method:adapt}), we crop the original FFHQ~\cite{karras2019style} data with a similar field of view to our synthetic data and resize the images to $512x512$ resolution. The FFHQ dataset contains 70k images, which are split into 60k/10k for training/validation. We use the training set in our adaptation and leave the validation set for evaluation. For temporal refinement (Section~\ref{sec:method:video}), we use the TalkingHead-1kH dataset~\cite{wang2021facevid2vid}, which consists of about 500k face video clips. We randomly sample $4$ consecutive frames from a clip during each iteration to train the temporal network.

For both our syn2real and temporal networks ($f_{\sbar{A}}, f_{\dbar{A}}, f_{\dbar{N}}$), we use a ResNet with $2$ downsample layers, $9$ residual blocks, and $2$ upsampling layers. The channel size for the first layer is $32$ and is doubled after each downsampling. We train our networks using a learning rate of 0.0001. For each of the three stages (synthetic, syn2real, and temporal), we train for 50k iterations using a batch size of $64$ on $8$ NVIDIA V100 GPUs. All of our networks are trained to produce an image of 512 resolution. Our end-to-end process runs in 65ms on an NVIDIA A6000 GPU. 

\section{Experiments} \label{sec:exp}

We compare our method with SIPR-W~\cite{wang2020single}, TR~\cite{pandey2021total}, and NVPR~\cite{zhang2021neuralvideo}. Since none of these methods releases their code or models, we requested the authors to apply their models to our inputs and provide the results to us. Note that for NVPR, the authors did not return the results for the synthetic test set, so we only compare with NVPR on the real dataset.
In the following, we first show comparisons on our synthetic test set (Section~\ref{sec:exp:syn}). We then show comparisons on a real in-the-wild face dataset using FFHQ (Section~\ref{sec:exp:real}). Next, we perform ablation studies on our network design choices (Section~\ref{sec:exp:ablation}). Finally, we show additional features available in our framework, including controlling glasses glares and relighting portrait videos with temporal consistency (Section~\ref{sec:exp:video}).

\subsection{Comparisons on the Synthetic Dataset} \label{sec:exp:syn}
\begin{figure}[t]
\centering
\includegraphics[width=\linewidth]{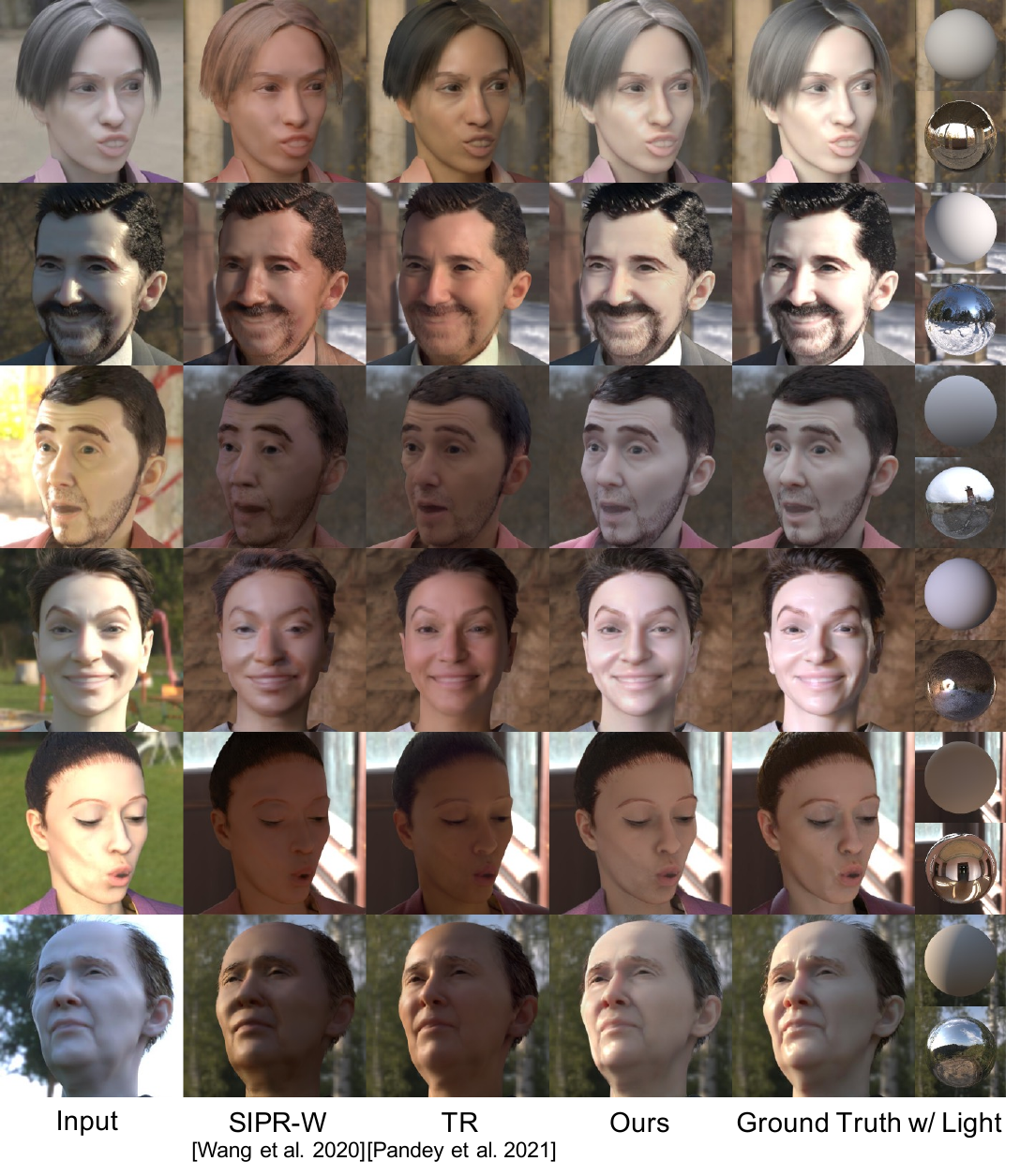}
    \vspace{-0.7cm}
  \caption{Comparisons with state-of-the-art methods on our synthetic dataset. SIPR-W~\cite{wang2020single} and TR~\cite{pandey2021total} tends to change skin tone and hair color. Ours can predict the relit results closer to the ground truth and preserve the identities, skin tone, hair color, and the facial details.}
\label{fig:comp_syn}
\end{figure}

\begin{figure*}[t!]
\centering
\includegraphics[width=0.95\linewidth]{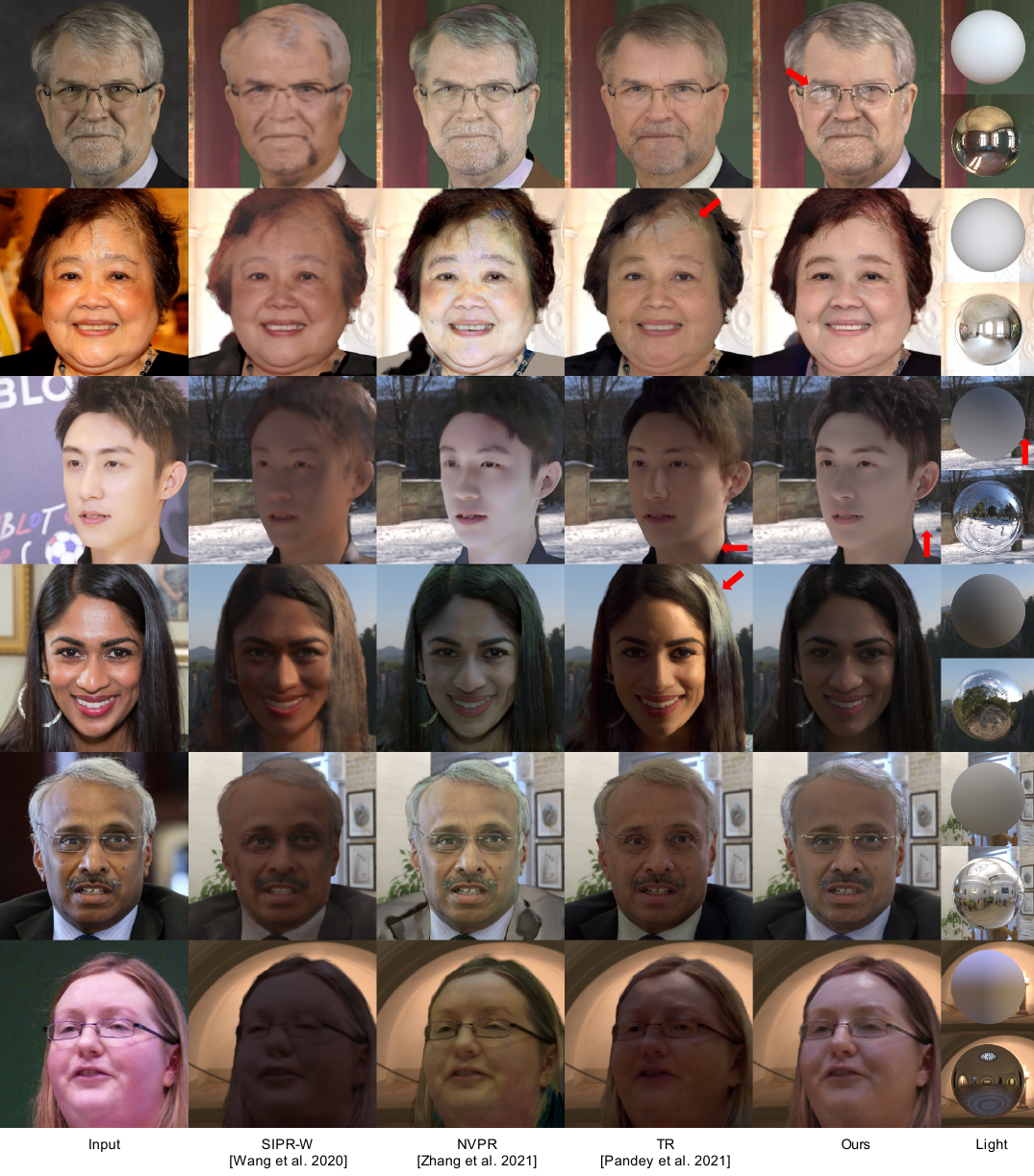}
    \vspace{-0.3cm}
  \caption{Comparisons with state-of-the-art methods on FFHQ evaluation dataset for real portraits. SIPR-W~\cite{wang2020single} cannot handle in-the-wild images due to the models purely trained with synthetic data. NVPR~\cite{zhang2021neuralvideo} generates unnatural appearance and artifacts on cloth regions. TR~\cite{pandey2021total} is able to generate plausible relit images, but exhibits artifacts on hair (second and fourth rows), inconsistent lighting directions with the reference diffuse sphere (third row), and lack of facial details (sixth row). Our results can synthesize plausible glares on eyeglasses (first row), keep facial and hair details, and generate relit portraits consistent with the target illumination conditions.   }
\label{fig:comp_real}
\end{figure*}

\begin{figure*}[t]
\centering
\includegraphics[width=\linewidth]{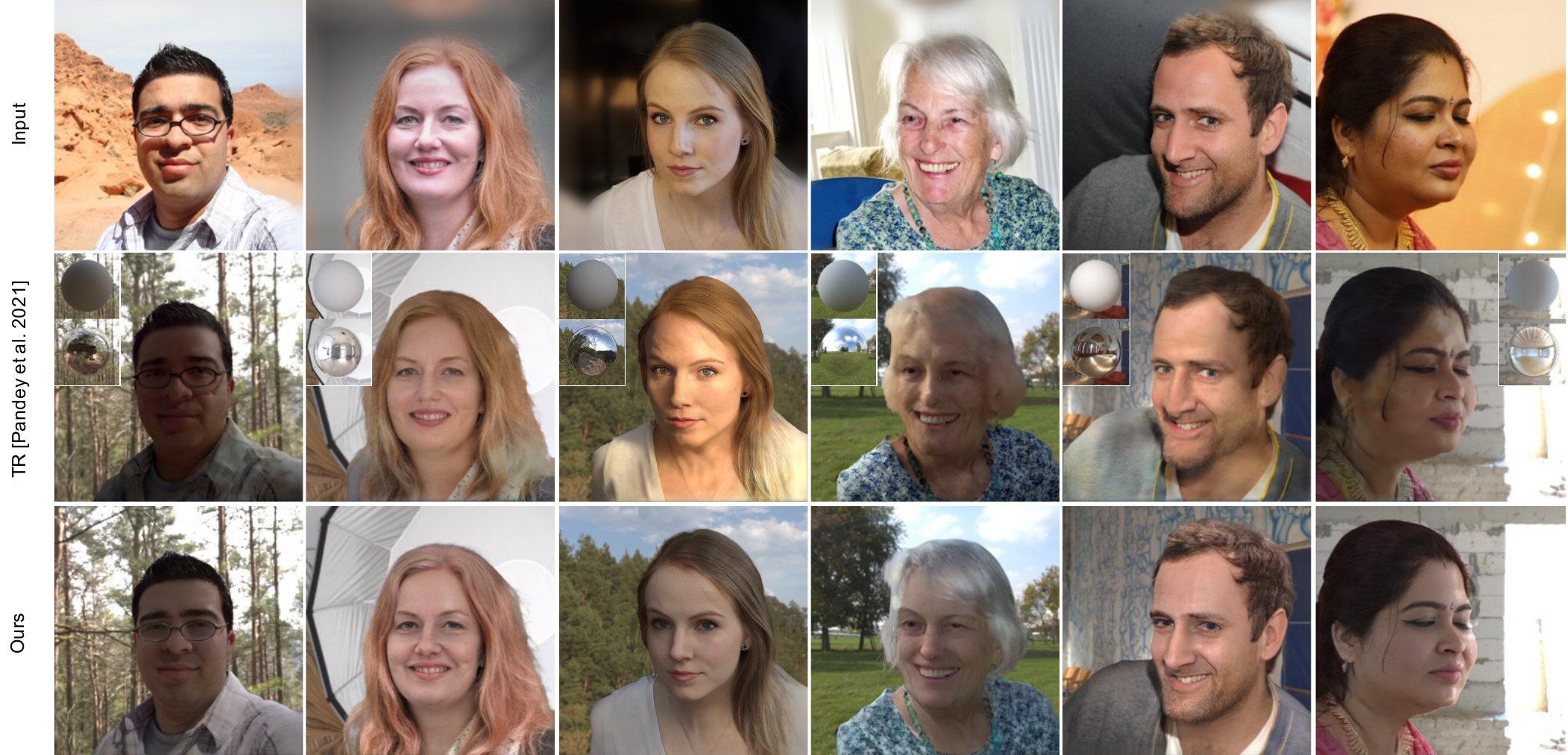}
    \vspace{-0.5cm}
  \caption{Comparisons with state-of-the-art methods on FFHQ evaluation dataset for \textit{larger} (including shoulder) real portraits. TR~\cite{pandey2021total} tends to generate blurred details for hair regions and shows artifacts on clothes. TR also exhibits the relit face that is inconsistent with the given lighting, compared to the diffuse sphere reference (third column). Ours can remove unwanted cast shadows from eyeglasses (first column), keep details and colors on clothes and hair, and overall consistent lighting with the diffuse sphere reference.}
\label{fig:comp_real_large}
\end{figure*}
Our synthetic test set consists of $25$ identities that are never seen during training. We use all the available expressions for each identity and randomly pair each of them with two environment maps to form a ground-truth pair. This results in $477$ pairs of images in total.

\subsubsection{Quantitative Results}
\begin{table}[!bt]
\begin{center}
\caption{Quantitative evaluations on the test set of our synthetic dataset.}
\vspace{-0.2cm}
\label{tab:quan_syn}
\small
\begin{tabular}{lcccc}
\toprule
& MAE $\downarrow$ & MSE $\downarrow$ & SSIM $\uparrow$ & LPIPS $\downarrow$ \\ \cmidrule(r){2-5}
SIPR-W~\cite{wang2020single} & 0.2506 & 0.1159 & 0.4406 & 0.3633 \\
TR~\cite{pandey2021total}    & 0.2328 & 0.1083 & 0.4851 & 0.3465 \\
Ours& \textbf{0.1967} & \textbf{0.0975} & \textbf{0.5272} & \textbf{0.2891} \\

\bottomrule
\end{tabular}
\end{center}
\vspace{-0.2cm}
\end{table}
We compare the mean absolute error (MAE), the mean squared error (MSE), the structural similarity index measure (SSIM)~\cite{wang2004image}, and the Learned Perceptual Image Patch Similarity (LPIPS)~\cite{zhang2018unreasonable} between the relit images and the ground truths for each method. The comparison is shown in Table~\ref{tab:quan_syn}. \YY{Note that all the baselines and our method suffer from the similar domain gap, i.e., trained on real data but evaluated on synthetic data.} Our method achieves the best results on all metrics, outperforming all other baselines by a large margin.

\subsubsection{Qualitative Results.}
The qualitative comparisons are shown in Figure~\ref{fig:comp_syn}. Compared with other methods, the results generated by our approach are more similar to the ground truth in terms of both lighting accuracy and color tones. The results by SIPR-W have multiple artifacts, especially on the hair and clothes. The results by TR have fewer artifacts but sometimes change the person's identity or skin colors. This may be expected since TR is never trained on synthetic data. Hence, we compare the performance of the challenging in-the-wild images next.

\subsection{Comparisons on the Real Dataset} \label{sec:exp:real}
We compare the performance of the competing methods on real images in the FFHQ dataset. We pick $116$ images from the FFHQ test set and randomly pair each of them with $4$ different environment maps, resulting in $464$ images in total. The images are chosen with gender, age, and ethnic diversities in mind.

\begin{table}[!bt]
\begin{center}
\caption{Quantitative evaluations on the test set of FFHQ.}
\vspace{-0.2cm}
\label{tab:ffhq}
\begin{tabular}{lcccc}
\toprule
& Lighting $\downarrow$ & FID $\downarrow$ & Identity $\uparrow$ \\ \cmidrule(r){2-4}
SIPR-W~\cite{wang2020single} & - & 73.31 & 0.6522 \\
NVPR~\cite{zhang2021neuralvideo}  & - & 52.25 & \textbf{0.7461} \\
TR~\cite{pandey2021total} & 0.0699 & 43.50 & 0.6226 \\ 
Ours     & \textbf{0.0645} & \textbf{36.13} & 0.7413 \\ 
\bottomrule
\end{tabular}
\end{center}
\vspace{-0.2cm}
\end{table}

\subsubsection{Qualitative Results.}
In Figure~\ref{fig:comp_real}, we provide qualitative comparisons in a tighter crop, accommodating the preferred setting in SIPR-W and NVPR. SIPR-W has trouble handling the real-world data since the model is only trained with synthetic data. NVPR tends to predict unnatural illumination on the face regions and incorrect appearances on clothes regions. TR is able to generate photorealistic results but exhibits artifacts on hair regions (second and fourth rows) and loses facial details (sixth row). In addition, the illumination by TR is sometimes not aligned with the target. In the third row, the ground snow is supposed to illuminate the face from the bottom as shown in the diffuse sphere reference, while the TR output illuminates the face from the right side. Our method generates a natural appearance and consistent lighting conditions with the reference sphere and additionally adds glares on eyeglasses (first row). In Figure~\ref{fig:comp_real_large}, we compare our results with the TR method in larger crops. While TR can overall generate photorealistic results, it shows unnatural appearance and artifacts in the hair regions and an inconsistent lighting output (third column). It also shows blurry results in the hair region (fourth column). Our method can robustly handle the wide variety of hair colors and clothes types thanks to the variations in our synthetic data and in-the-wild data used in the synthetic-to-real adaptation.  

\subsubsection{Quantitative Results.}
Since we do not have ground truths for quantitative comparisons, we adopt the following metrics. We measure the resulting quality of each method in three different aspects: lighting accuracy, photorealism, and identity preservation.
\vspace{-1mm}
\paragraph{Lighting recovery.}
\begin{figure*}[t]
\centering
\includegraphics[width=\linewidth]{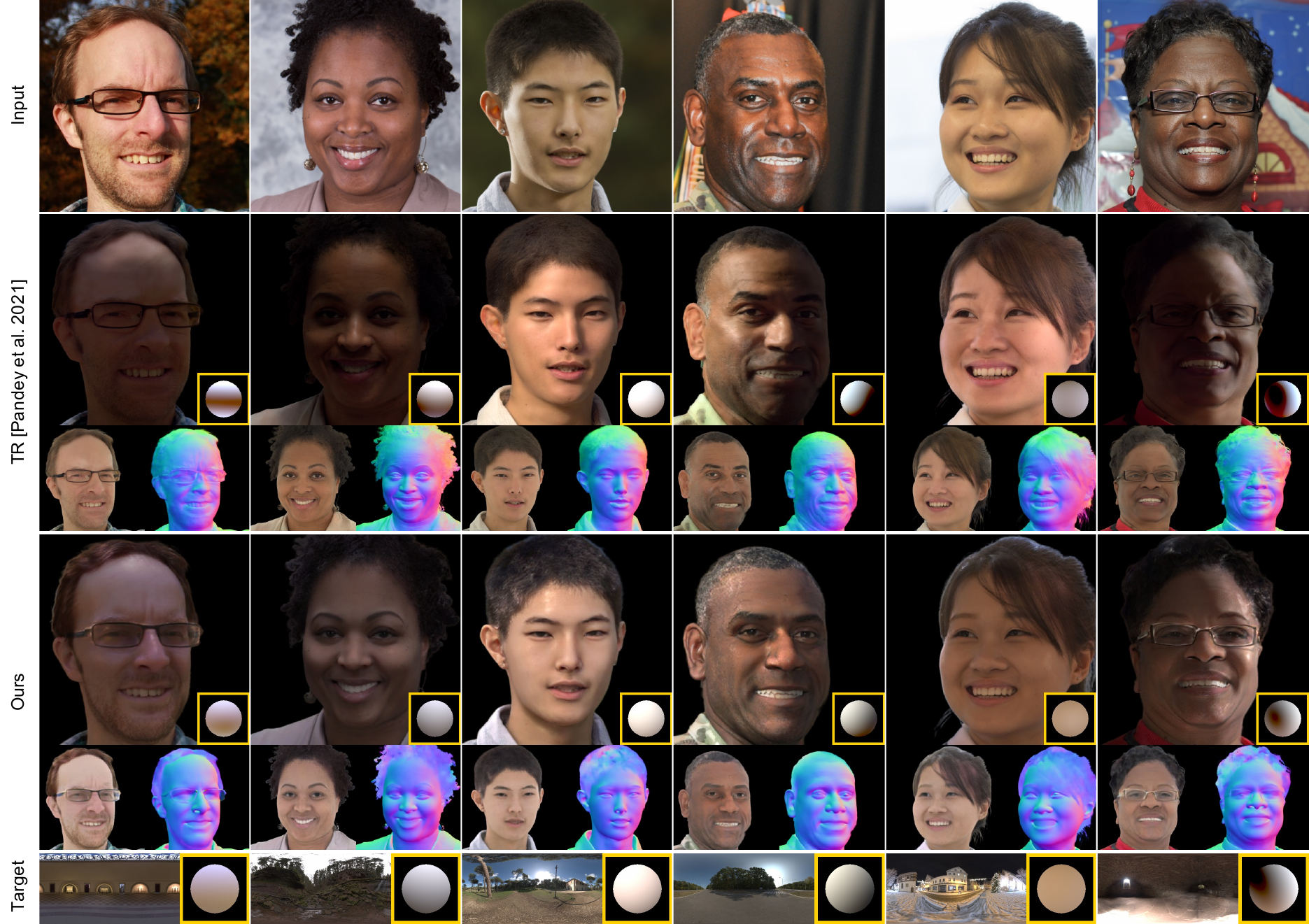}
    \vspace{-0.5cm}
  \caption{Comparison of lighting recovery from relit images on the FFHQ dataset. 
We compare TR (second row) and our method (third row) on the consistency of the lighting information recovered from the relit portrait via the second-order spherical harmonic (SH) lighting~\cite{ramamoorthi2001efficient} compared to the ground truth environment (bottom row). Given the relit image (top of the row), albedo (bottom left of the row), and normal map (bottom right of the row) generated from the input image (top row), we estimate the second-order SH coefficients by solving a linear system using foreground pixels. We visualize the recovered SH lighting condition on the diffuse ball rendering in the inset highlighted in yellow. The inset diffuse ball renderings show our method has more consistent lighting than TR compared to the ground truth diffuse ball rendering (bottom row).}
\label{fig:light_recovery}
\end{figure*}
To evaluate how accurately the relit results translate the lighting information from the input lighting environment, we estimate the lighting from relit images and compare it with the target lighting. In particular, we use the relit portrait, along with the estimated albedo and normal maps, to compute the second-order spherical harmonic (SH) coefficients~\cite{ramamoorthi2001efficient} by solving a least square problem on visible skin pixels. We use face \textit{skin} and \textit{nose} masks predicted from a PyTorch implementation of a face parsing network\footnote{\url{https://github.com/zllrunning/face-parsing.PyTorch}} based on BiSeNet~\cite{yu2018bisenet} to determine skin pixels. This is done only for TR~\cite{pandey2021total} and our method since both methods generate albedo and normal maps as side products. We reconstruct a sphere rendering from estimated SH coefficients and apply a global intensity multiplier to compensate for a global exposure difference between the estimated and target illumination. To be specific, we compute the average intensity of the sphere rendering from the estimated SH coefficients to derive the global scalar for matching the average intensity of the sphere rendering from the target SH coefficients. We compute the L1 pixel difference between the reconstructed and target sphere renderings to measure the accuracy. We computed this error in the pixel space instead of the coefficient space since the coefficient space has ambiguity (i.e., multiple coefficient combinations can give rise to the same reconstruction). The comparisons are shown in Figure~\ref{fig:light_recovery} and Table~\ref{tab:ffhq}. Our results demonstrate better numerical accuracy and are closer to the ground truth sphere rendering. \YY{Although the intermediate normal predictions from TR might contain more high-frequency details than ours, it sometimes predicts incorrect geometry, such as the eyeglasses shadow in the cheek region (first column in Figure~\ref{fig:light_recovery}). When the entire framework is trained end-to-end, the final rendering network is able to compensate for small differences or errors from the normal prediction.}
\vspace{-1mm}
\paragraph{Synthesis quality.} To compare the photorealism of the resulting quality, we compute Fréchet Inception Distance (FID)~\cite{heusel2017gans,Seitzer2020FID} scores between the input FFHQ test set and the relit results from each method. For fair comparisons, we mask out the background of each image and only consider the foreground region. From Table~\ref{tab:ffhq}, it can be seen that our method has the lowest FID score, which demonstrates its ability to synthesize realistic portrait images.
\vspace{-1mm}
\paragraph{Identity similarity.}
We evaluate the identity consistency before and after relighting, as shown in Table~\ref{tab:ffhq}. To do so, we use a face recognition network~\cite{an2020partical_fc} (different from the one that we use in our training loss) to extract face embeddings before and after relighting of foreground images for each method and compute the similarity score between these two embeddings. We found that NVPR achieves a slightly higher identity similarity score, but our score is very close to theirs, and our FID score is much better. Moreover, from Figure~\ref{fig:comp_real} it can be seen that their method shows unnatural facial appearance that is not consistent with the target illumination (fourth and sixth rows). 

\begin{table}[!bt]
\begin{center}
\caption{User study on the FFHQ test set. For each comparison, we ask the workers to choose the better image according to the criterion on the left. The preference rates where workers prefer our image than the other baselines are summarized below. For all metrics, the preference rates are greater than 0.5, meaning the workers prefer our results more.}
\vspace{-0.2cm}
\label{tab:user_ffhq}
\small
\begin{tabular}{lccc} \toprule
        & Ours vs.\ SIPR-W & Ours vs.\ NVPR & Ours vs.\ TR \\ \cmidrule(r){2-4}
Consistent Lighting & 0.5453 & 0.5934 & 0.5474 \\ 
Facial Details   &  0.8527 & 0.7601 & 0.5409 \\
Similar Identity & 0.8886 & 0.8218 & 0.5230  \\ 
\bottomrule
\end{tabular}
\end{center}
\vspace{-0.2cm}
\end{table}

\vspace{-1mm}
\paragraph{User study.}
Finally, we perform a user study by asking Amazon Mechanical Turk workers to answer three binary questions: given an input image and an environment map, which relit image \textit{1) has more consistent lighting with the environment map? 2) maintains better facial details? 3) keeps the original identity better?} between our relighting result and one of the baselines. The orders of the methods are randomized for fair comparisons. To make the workers better understand the concept of consistent lighting, we also show the relit balls similar to the ones shown in Figure~\ref{fig:comp_real}. \YY{We split the 464 images into 4 partitions and ask each worker to evaluate a partition on a single question.} Each comparison is evaluated by $3$ different workers, resulting in $1392$ comparisons for each \YY{question}. \YY{In total, 4 (partitions) x 3 (questions) x 3 (comparisons) = 36 workers participated who have never seen the questions.} The evaluation results are shown in Table~\ref{tab:user_ffhq}. It can be seen that workers prefer our results more compared to all baselines using all different metrics. More details on the user study interface can be found in Appendix~\ref{sec_a:user_study}

\subsection{Ablation Studies} \label{sec:exp:ablation}
\begin{table}[!bt]
\begin{center}
\caption{Ablation study on baseline adaptation methods with real FFHQ test set. All the results are computed on foreground images. There are trade-offs between identity consistency, synthesis quality (FID) and OLAT consistency. The model can report a slightly higher FID score or the identity similarity when it is trained \textit{without} relative or OLAT consistency loss at the cost of the much worse lighting consistency, which is less suitable for portrait relighting. Our full model achieves the \textbf{best} performance in OLAT consistency while maintaining the similar scores (\underline{second best}) in the other metrics, which leads to the best lighting control while preserving the portrait identities with high photorealism.
}
\vspace{-0.2cm}
\label{tab:ablation}
\begin{tabular}{lccc}
\toprule
                  & OLAT Cons. $\downarrow$ & FID $\downarrow$ & Identity $\uparrow$ \\ \cmidrule(r){2-4} 
w/o improved render net  & 0.0970           & 54.08          & 0.7237   \\ 
\cmidrule(r){1-4}
w/o adaptation               & 0.0417          & 58.61          &  0.6402  \\ 
Finetune only                & 0.2284          & 69.24          &  0.6932   \\ 
Direct syn2real              & 0.0267          & 43.14          &  0.7206   \\ \cmidrule(r){1-4}
w/o Relative consistency loss    & 0.0746          & 36.90          & \textbf{0.7612} \\ 
w/o OLAT consistency loss    &\underline{0.0205}          & \textbf{35.87} &  0.7405 \\ \cmidrule(r){1-4}
Our full model               & \textbf{0.0166} & \underline{36.13} & \underline{0.7413} \\ 
\bottomrule
\end{tabular}
\end{center}
\vspace{-0.2cm}
\end{table}

\begin{figure*}[t]
\centering
\includegraphics[width=\linewidth,trim=0 7pt 0 0, clip]{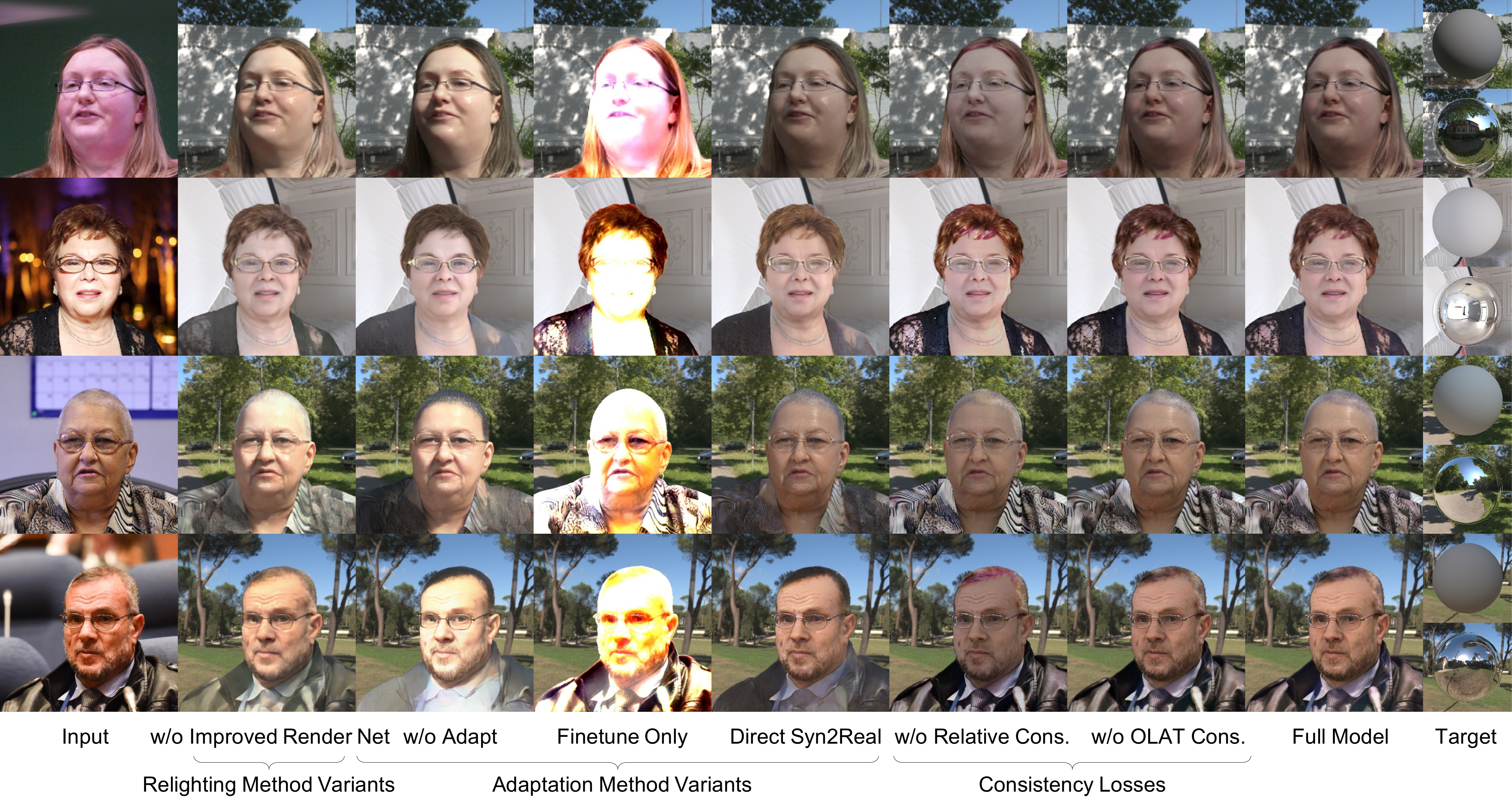}
    \vspace{-0.7cm}
  \caption{Ablation comparisons on our adaptation choices. For \textit{w/o adaptation}, the model cannot perform well on real data and cannot keep the facial details such as hair color and beards. For \textit{w/o Improved Render Net}, \textit{Finetune Only}, and \textit{Direct Syn2Real}, they give too much freedom to the network which also leads to degenerated results due to overfitting, including out-of-control relighting results and losing facial or clothing details. With our proposed adaptation method, \textit{w/o Relative consistency loss} cannot synthesize minor facial highlights and might change hair or face colors. For \textit{w/o OLAT consistency loss}, the overall results are similar to our full model, but the lighting control is slightly worse and the output exhibits artifacts on the hairlines (first, second and fourth rows). }
  \vspace{-0.3cm}
\label{fig:ablation}
\end{figure*}
We compare different baselines for our relighting model design choices described in Section~\ref{sec:method:syn} and Section~\ref{sec:method:adapt}. We compute the same metrics as used in Section~\ref{sec:exp:syn} for synthetic test set and in Section~\ref{sec:exp:real} for the real test set in Table~\ref{tab:ablation}. In addition, we also compute the OLAT consistency error to evaluate the lighting control of our relighting performance. The computing procedure is similar to our OLAT lighting consistency loss (Equation~\ref{eqn:olat_loss}). The qualitative results of each ablative baseline are shown in Figure~\ref{fig:ablation} on the FFHQ real test set. We describe each ablative baseline as follows:
\vspace{-2mm}
\paragraph{w/o improved render network.} Instead of using the proposed rendering decomposition and delta relit image prediction, we directly use a blackbox rendering network in TR~\cite{pandey2021total} to predict the relit images.
\vspace{-2mm}
\paragraph{w/o adaptation.} We only use the proposed improved rendering network (Section~\ref{sec:method:syn}) without applying the syn2real adaptation (Section~\ref{sec:method:adapt}).
\vspace{-2mm}
\paragraph{Finetune only.} We apply all the syn2real adaptation losses mentioned in Section~\ref{sec:method:adapt} to finetune all the networks of our base model (Section~\ref{sec:method:syn}) after training with synthetic data.
\vspace{-2mm}
\paragraph{Direct syn2real on relit image.} Instead of learning an albedo residual $\delta{\tilde{A}}$ as mentioned in Section~\ref{sec:method:adapt}, we learn a module to predict the residual for the output relit image while keeping the weights for the other modules in the base model fixed.
\vspace{-2mm}
\paragraph{w/o relative consistency loss.} We keep the weights of our base model unchanged and only optimize the weights for the albedo residual network with all losses for adaptation except the relative consistency loss.
\vspace{-2mm}
\paragraph{w/o OLAT consistency loss.} Similar to above, we only optimize for the albedo residual network with all adaptation losses except the OLAT consistency loss.
\vspace{-2mm}
\paragraph{Full adaptation.} Our full adaptation method optimizes the albedo residual network with all the adaptation losses.

As shown in Figure~\ref{fig:ablation}, \textit{w/o adaptation} cannot generalize to real portrait images well. It predicts incorrect appearances on hair and clothing regions. For \textit{w/o improved render net}, \textit{Finetune only}, and \textit{Direct syn2real}, they give too much freedom to the adaption network during training, which leads to incorrect relighting results due to overfitting. In particular, the overall appearances do not reflect the input environment maps for \textit{Finetune only}. The facial and clothing details are not preserved well in \textit{w/o improved render net}. For \textit{Direct syn2real}, although the relighting results look plausible, it loses the details of clothing regions. On the other hand, \textit{w/o relative consistency loss} loses minor relighting details such as the specular highlights on the skin (third row) and might also change hair or face colors. For \textit{w/o OLAT consistency loss}, the relit image exhibits artifacts around the hairline (first, second and fourth columns). 

We also would like to note there are trade-offs between identity similarity, synthesis quality (FID), and lighting control. One possible way for the model to optimize for the identity similarity and the FID score is to try not to change image content as much as possible at the cost of \YY{violating} lighting consistency. As seen in Table~\ref{tab:ablation}, \textit{w/o OLAT consistency loss} and \textit{w/o relative consistency loss} achieve the slightly better FID and identity scores but also show much worse lighting consistency. This is reinforced in Figure~\ref{fig:ablation} that without the lighting consistency losses, the outputs exhibit unnatural results and artifacts less suitable for relighting. We choose our full model with the \textbf{best} performance on OLAT consistency while keeping the identity similarity and FID scores as the \underline{second best}. This leads to the best lighting control on the relit images with the least artifacts while largely preserving the input identities with high photorealism.

\subsection{Video Relighting} \label{sec:exp:video}
Finally, we show results using our video relighting framework. 
\begin{figure}[t!]
    \centering
    \href{https://research.nvidia.com/labs/dir/lumos/images/glare/output2.gif}{\includegraphics[width=.96\columnwidth]{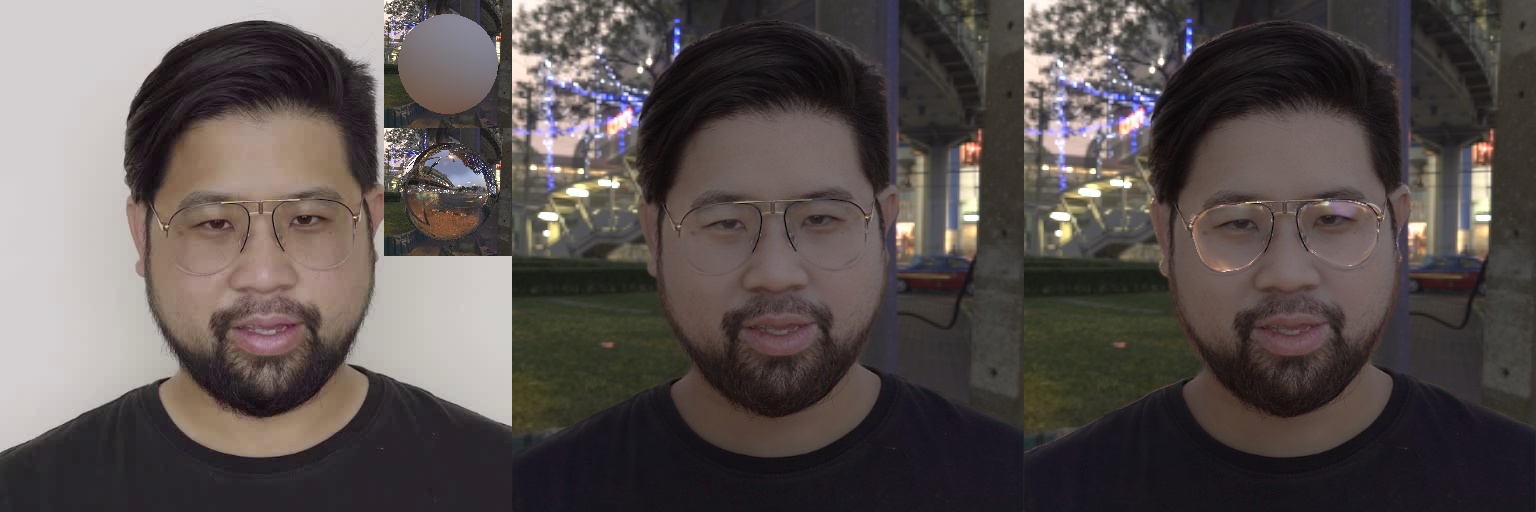}}
    \href{https://research.nvidia.com/labs/dir/lumos/images/glare/output3.gif}{\includegraphics[width=.96\columnwidth]{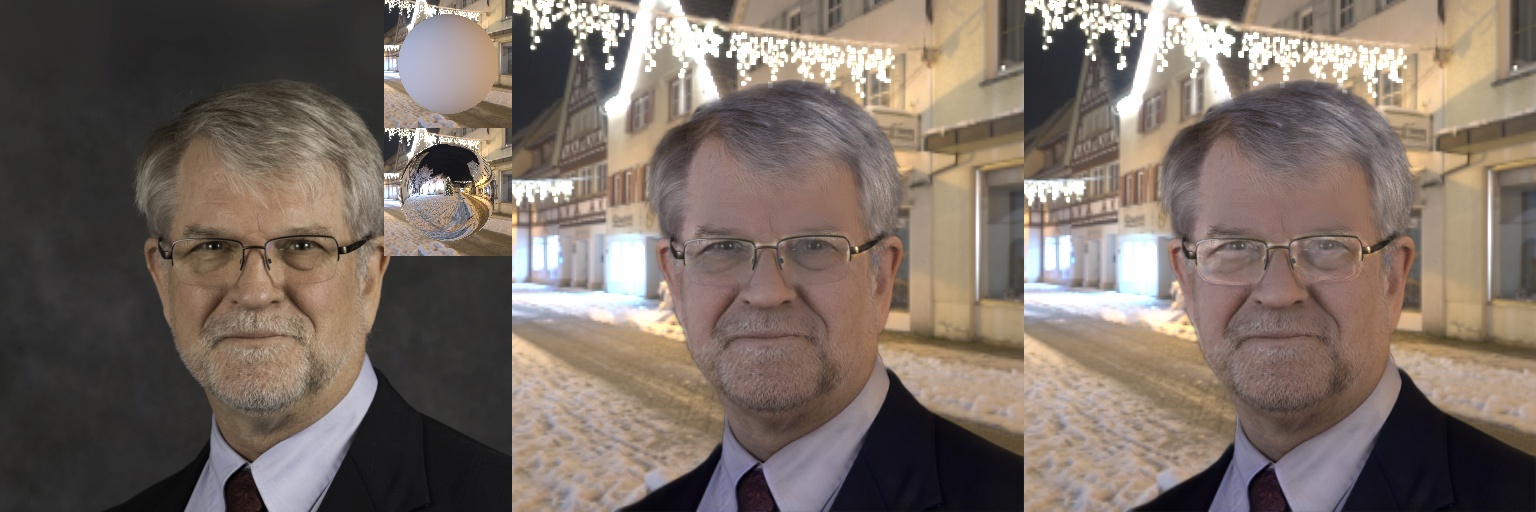}}
    \href{https://research.nvidia.com/labs/dir/lumos/images/glare/output4.gif}{\includegraphics[width=.96\columnwidth]{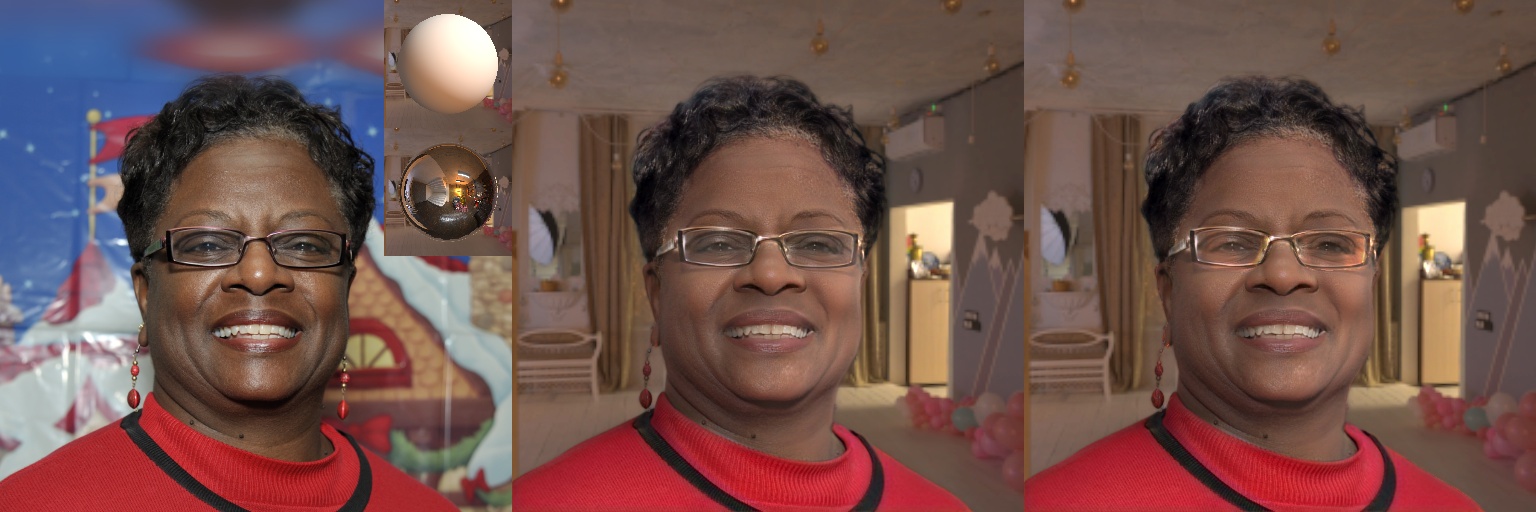}}
    \put(-205,-10){Input}
    \put(-150,-10){w/o Glasses Glares}
    \put(-70,-10){w/ Glasses Glares}
    \vspace{-0.3cm}
    \caption{Example comparisons showing results without vs.\ with adding glares using our method. \YY{The diffuse and mirror balls are shown in the set.}
    \emph{\textbf{Please click each row to view the video.}}
    }
    \label{fig:glasses_glares}
\end{figure}
\vspace{-2mm}
\paragraph{Controllable glasses glares.} Thanks to our synthetic dataset and our new network architecture (Section~\ref{sec:method:syn}), we are able to control the glares on glasses in the relighting results. This is achieved by three of our design choices. First, we have both normals with and without lenses in our synthetic dataset, so we are able to supervise our model on both normals. Second, we modify our geometry network and specular network to consider normals with lenses, including an additional light map with a very high Phong exponent. Finally, we directly use the light maps to first generate a coarse relit image and predict the residual for the output (Equation~\ref{eqn:init_render}) instead of using a neural network as a black box. Therefore, differences in light maps are directly transferable to the final output.

Example results are shown in Figure~\ref{fig:glasses_glares} in the embedded videos. \YY{Note that the glares are caused by the reflections from the rotated environment map.}

\begin{figure}[t]
\centering
    \href{http://research.nvidia.com/labs/dir/lumos/images/paper_videos/temporal_consistency/02.mp4}{\includegraphics[width=.95\columnwidth]{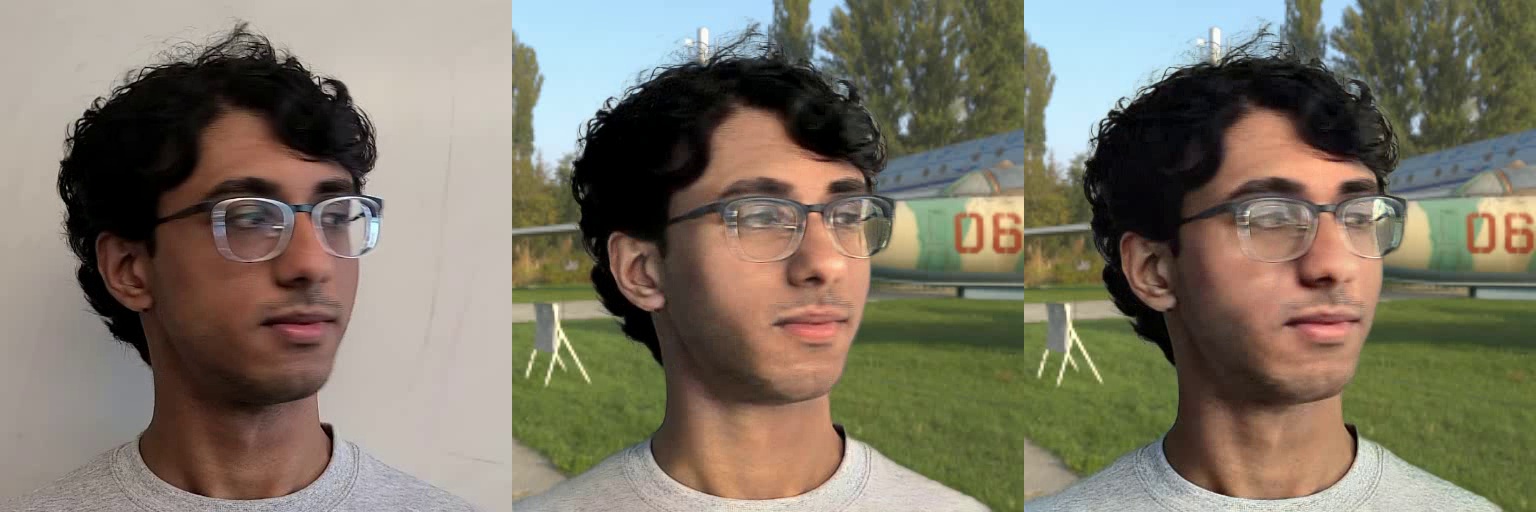}}
    \put(-200,-10){Input}
    \put(-140,-10){w/o Temporal}
    \put(-60,-10){w/ Temporal}
    \vspace{-0.3cm}
  \caption{Comparisons between with and without applying our temporal refinement. 
  \emph{\textbf{Please click each row to view the video.}}
  }
  \vspace{-0.3cm}
\label{fig:comp_temporal}
\end{figure}

\begin{table}[!bt]
\begin{center}
\caption{Quantitative comparisons for temporal consistency on relit videos. For each relit video, we use optical flows to warp neighboring frames and compute the differences.}
\label{tab:temporal}
\begin{tabular}{lcc}
\toprule
            & MAE $\downarrow$ & MSE $\downarrow$  \\ \cmidrule(r){2-3}
Ours (w/o temporal) & 0.0511 & 0.0024 \\ 
NVPR & 0.0488 & 0.0031 \\ 
Ours (w/ temporal) & \textbf{0.0477} & \textbf{0.0020} \\ 
\bottomrule
\end{tabular}
\end{center}
\vspace{-0.4cm}
\end{table}
\vspace{-2mm}
\paragraph{Temporally-consistent video relighting.} We are able to generate temporally-consistent outputs by refining the intermediate normal and albedo maps based on previous frames (Section~\ref{sec:method:video}). An example comparison is shown in Figure~\ref{fig:comp_temporal} in the embedded video. Without the temporal refining components, the results exhibit flickering artifacts when the subject is making large motions like rotating the head abruptly. When utilizing our temporal refining networks, the results become much smoother and more realistic.

\begin{figure}[t]
\centering
    \href{https://research.nvidia.com/labs/dir/lumos/images/paper_videos/temporal/00.mp4}{\includegraphics[width=.95\columnwidth]{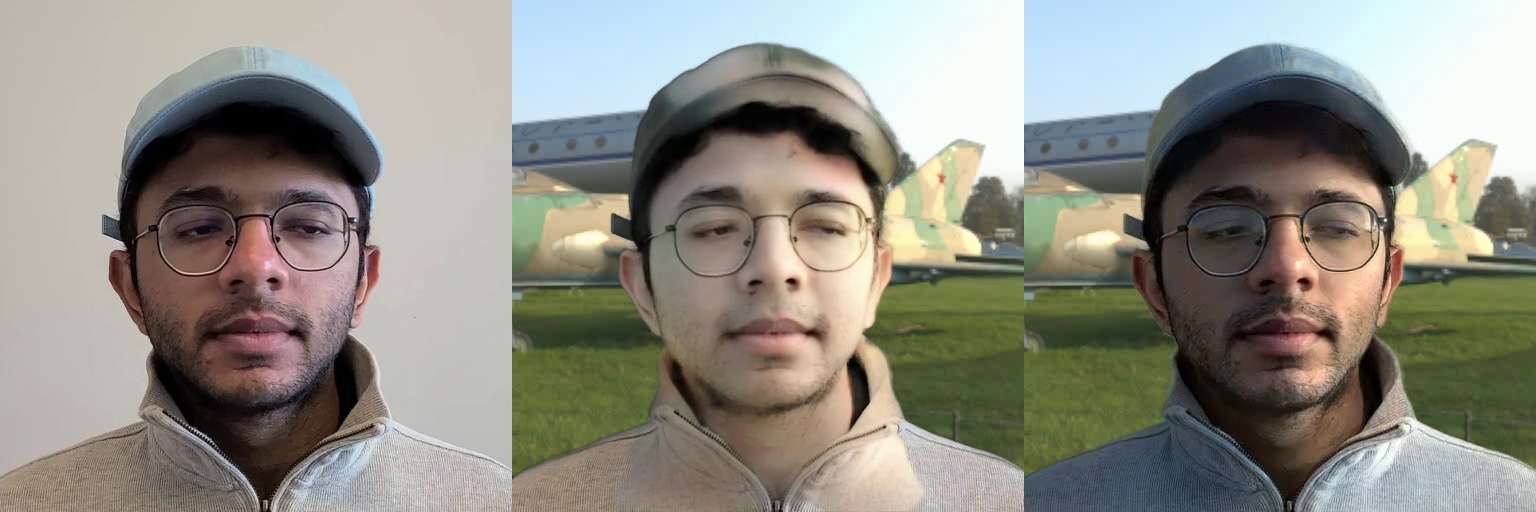}}
    \href{https://research.nvidia.com/labs/dir/lumos/images/paper_videos/temporal/01.mp4}{\includegraphics[width=.95\columnwidth]{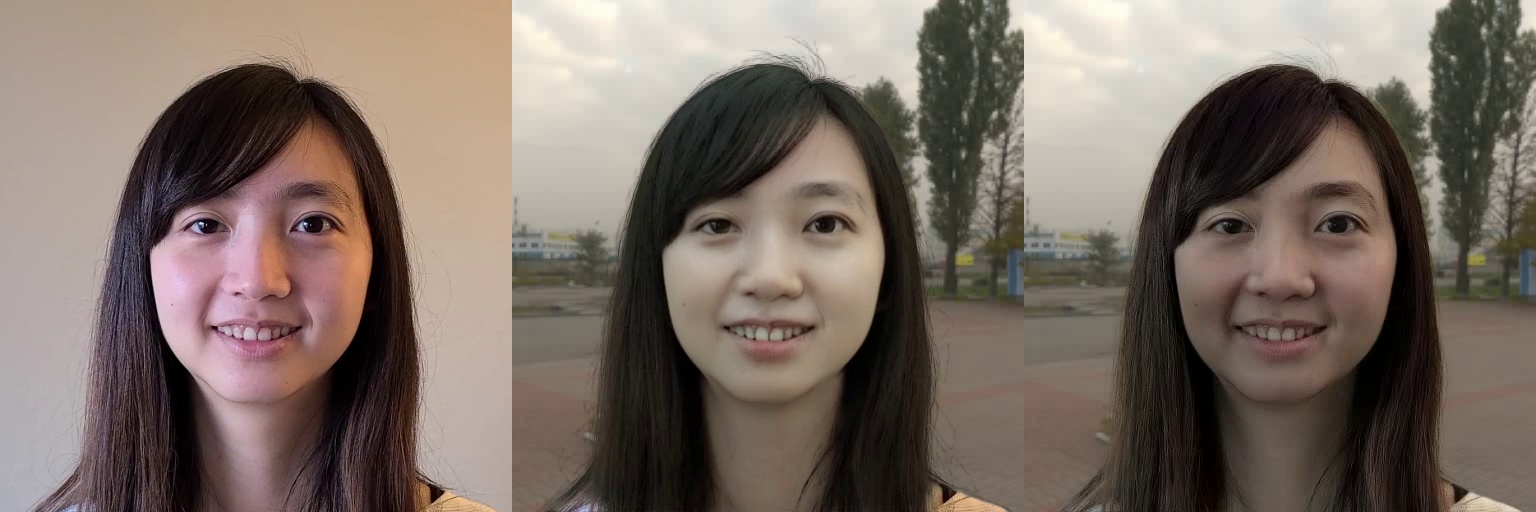}}
    \href{https://research.nvidia.com/labs/dir/lumos/images/paper_videos/temporal/02.mp4}{\includegraphics[width=.95\columnwidth]{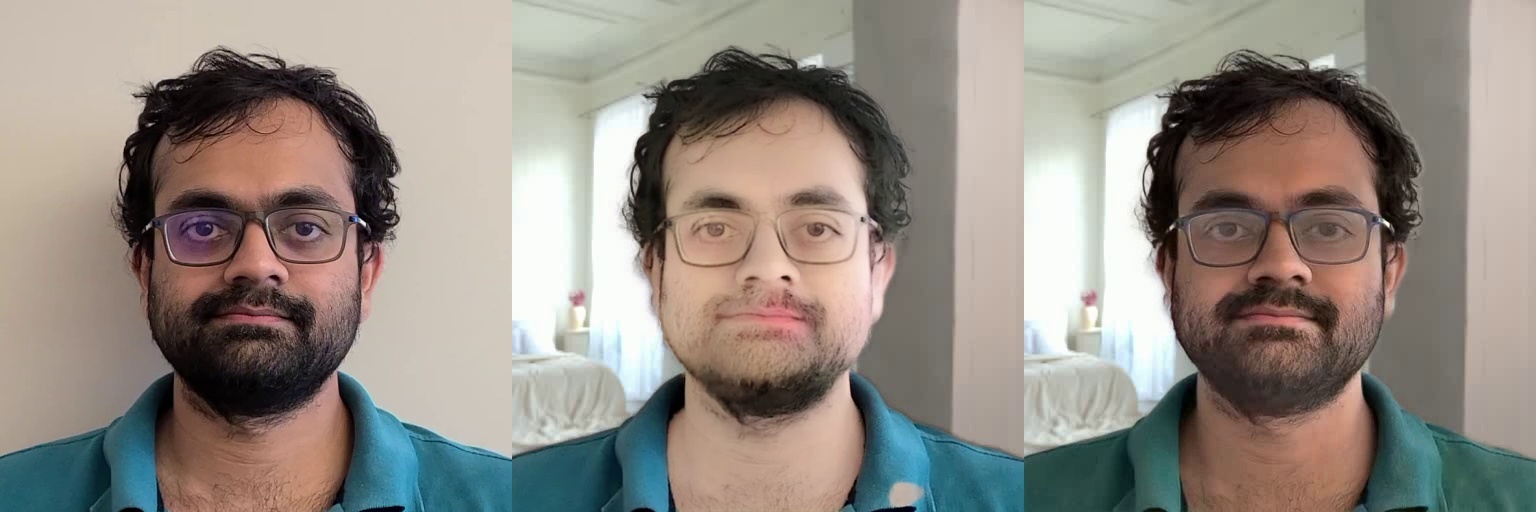}}
    \href{https://research.nvidia.com/labs/dir/lumos/images/paper_videos/temporal/03.mp4}{\includegraphics[width=.95\columnwidth]{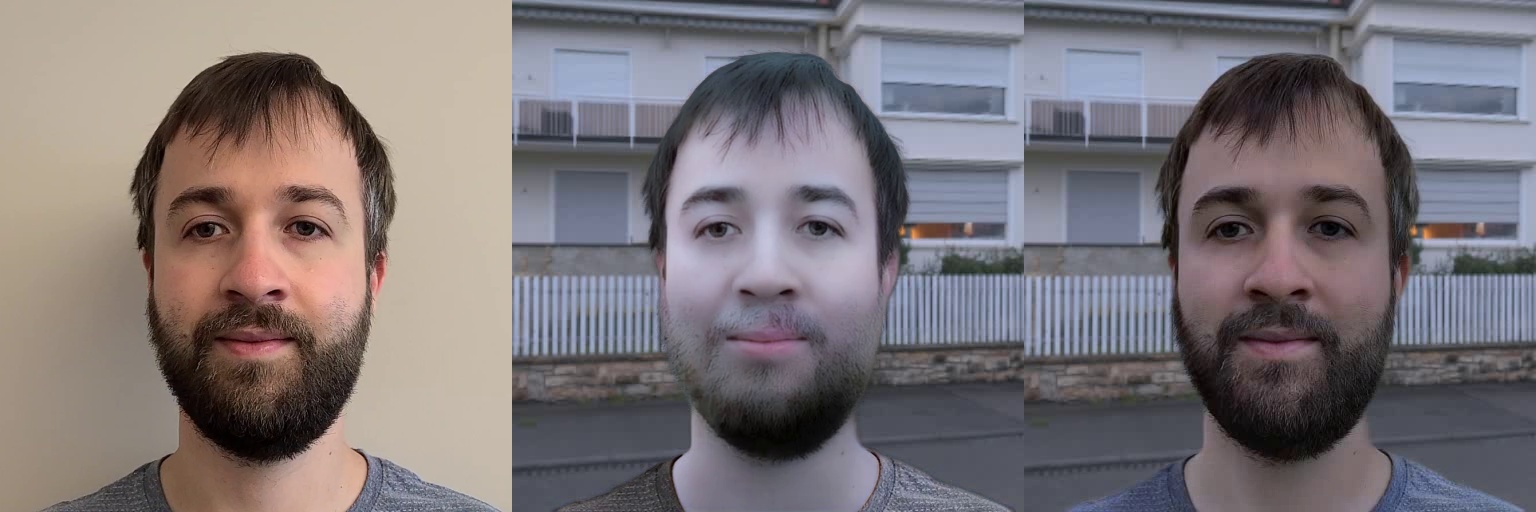}}
    \put(-200,-10){Input}
    \put(-160,-10){NVPR~\cite{zhang2021neuralvideo}}
    \put(-45,-10){Ours}
    \vspace{-0.3cm}
  \caption{Comparisons with the state-of-the-art method on video relighting.
  Note our results exhibit fewer flickering artifacts and are more temporally consistent. 
  \emph{\textbf{Please click each row to view the video.}}
  }
  \vspace{-0.5cm}
\label{fig:comp_video}
\end{figure}

We also compare our video relighting results with NVPR~\cite{zhang2021neuralvideo} in Figure~\ref{fig:comp_video} and Table~\ref{tab:temporal}. We evaluate temporal consistency on 9 different subjects with 6 selected environment maps (resulting in $54$ videos). We compute optical flows~\cite{flownet2-pytorch} on neighboring frames and warp the frame from $t$ to $t+1$ and compute the MAE and MSE between the warped frames and estimated frames. A lower error means the video contains fewer flickering artifacts. Ours performs better than NVPR in terms of temporal consistency and relighting quality.
\begin{figure}[t]
\centering
\includegraphics[width=\linewidth]{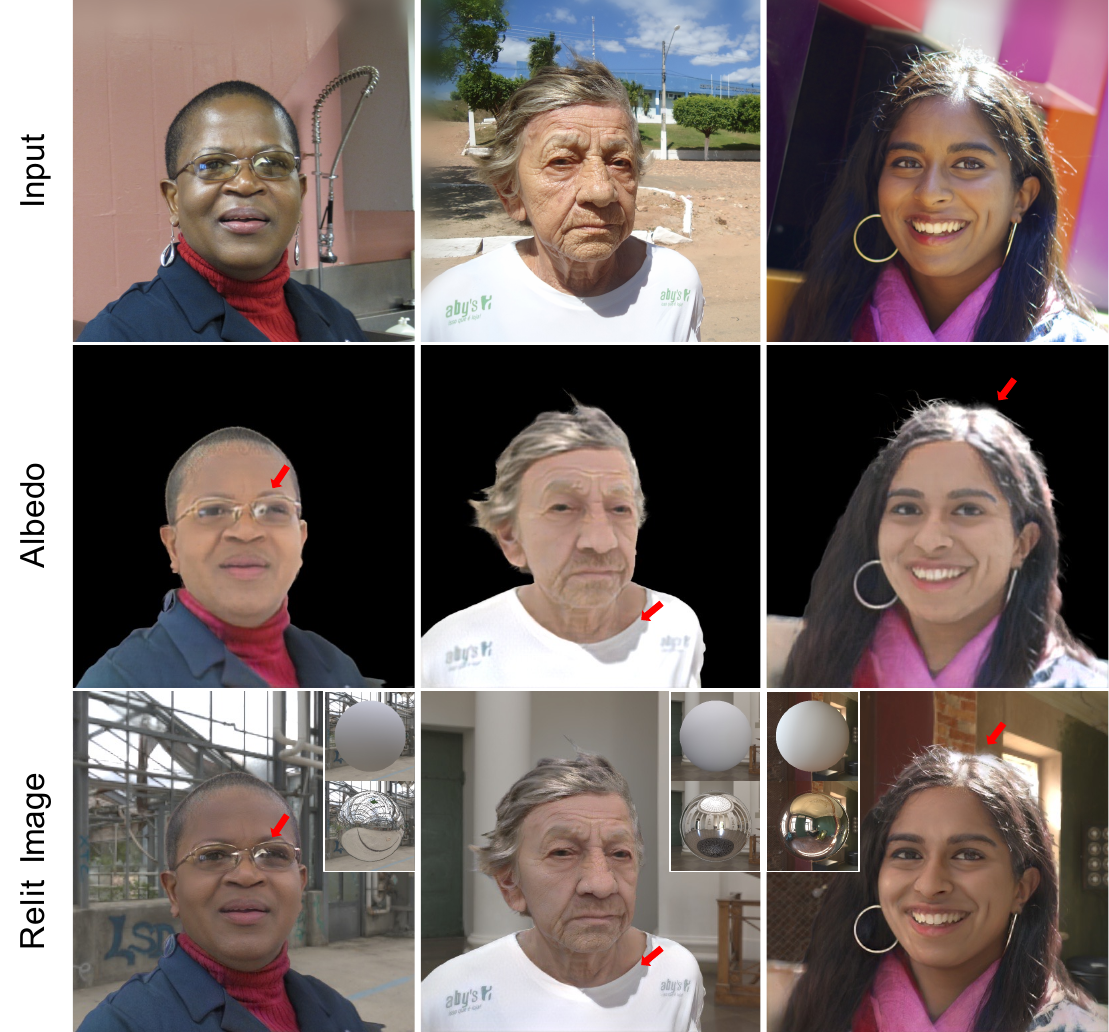}
    \vspace{-0.7cm}
  \caption{Examples of some failure cases. Our albedo estimation cannot remove strong highlights on eyeglasses and hair. The strong shadow cast on the clothes from the back of the subject is difficult to remove. The inaccurate albedo estimations lead to incorrect relit outputs.}
  \vspace{-0.5cm}
\label{fig:failure}
\end{figure}

\subsection{Limitations}

We demonstrate a few failure cases in Figure~\ref{fig:failure}.
Our method cannot predict the correct albedo for the overly exposed highlights on hair under a strong light source. It also cannot remove the reflection on the eyeglasses and the shadows cast by a direct light on the clothes. The inaccurate albedo predictions lead to incorrect relighting results. We could ameliorate this effect by first using an off-the-shelf method to remove these unwanted highlights or shadows and then applying our relighting model. For future work, we could explicitly model these strong shadows~\cite{zhang2020portrait} or reflections to learn better albedo and relighting predictions.
\section{Conclusion}
\label{sec:conclusion}
We proposed a new synthetic dataset for single image portrait relighting, which contains over 500 subjects of diverse ages, genders, and races with randomly assigned hairstyles, clothes, and accessories, rendered using a physically-based renderer. We showed that the single image and video portrait relighting tasks can be tackled without light stage data by training a base relighting model on the diverse synthetic dataset and later applying a synthetic-to-real domain adaptation using real in-the-wild portrait images and video. Self-supervised lighting consistency losses are used to ensure lighting consistency during the adaptation learning. Extensive experimental validations suggested that our relighting method outperforms the latest state-of-the-art single image or video portrait relighting methods in terms of lighting recovery, synthesis quality, and identity preservation. We believe our synthetic dataset for relighting and our method that does not require high-quality captured data fosters the portrait relighting research.

\begin{acks}
We thank Eugene Jeong, Miguel Guerrero, Simon Yuen, and Yeongho Seol for helping with generating the synthetic dataset. We also thank Manmohan Chandraker for insightful discussions and supports.
\end{acks}

\bibliographystyle{ACM-Reference-Format}
\bibliography{egbib}

\appendix
\section{Appendix}
\label{sec_a:dataset}
\subsection{Additional details on the synthetic dataset}

\begin{figure*}[t]
\centering
\includegraphics[width=\linewidth]{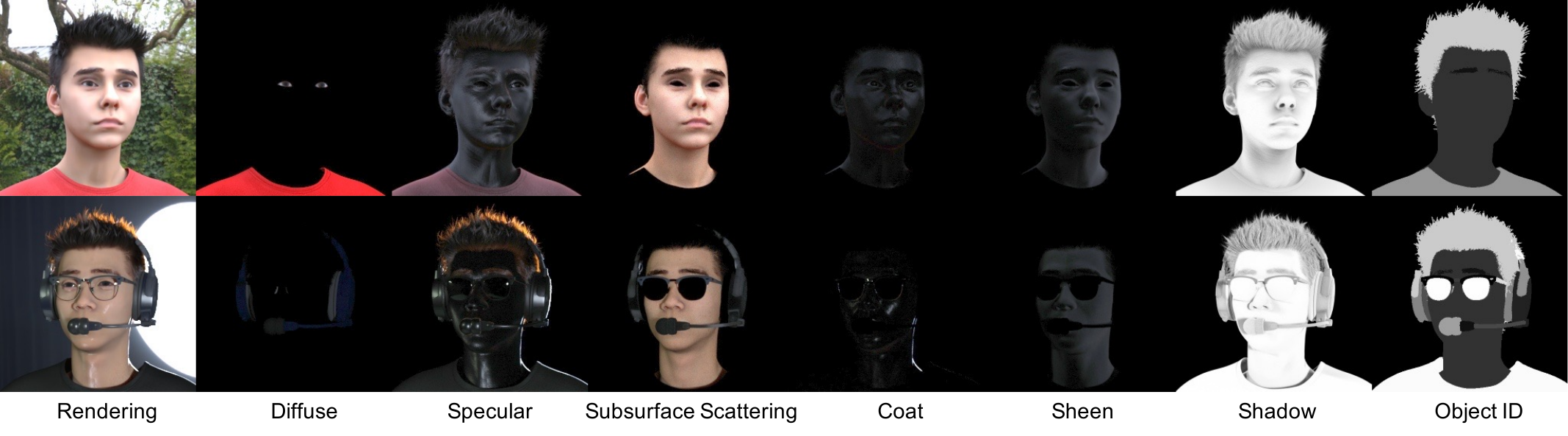}
    \vspace{-0.7cm}
  \caption{Example of additional rendering for implicit components. Diffuse, specular, subsurface scattering, coat and sheen are used in Arnold aiStandardSurface for face materials. We also explicitly rendered shadow maps and object ID maps which might be useful for future research.}
  \vspace{-0.1cm}
\label{fig:rendering_more}
\end{figure*}

\begin{figure*}[t]
\centering
\includegraphics[width=\linewidth]{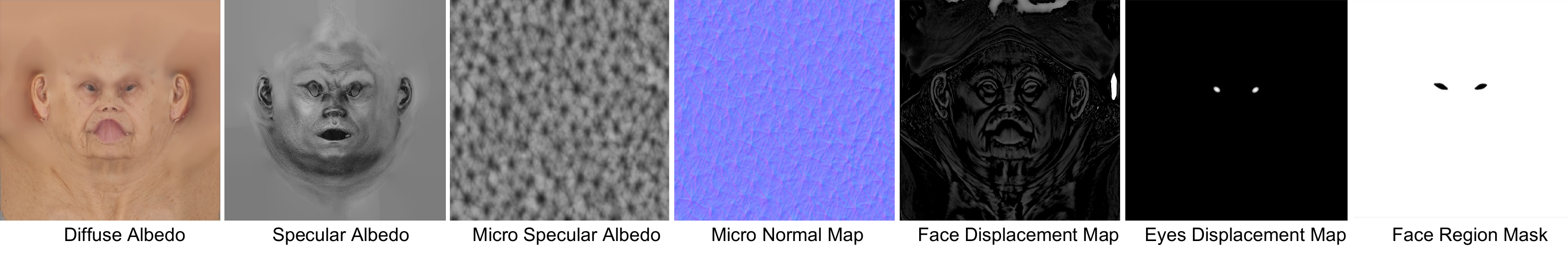}
    \vspace{-0.7cm}
  \caption{Example of spatially-varying maps used in Arnold \textit{aiStandardSurface} shader.}
  \vspace{-0.1cm}
\label{fig:shader_map}
\end{figure*}

\begin{figure*}[t]
\centering
\includegraphics[width=\linewidth]{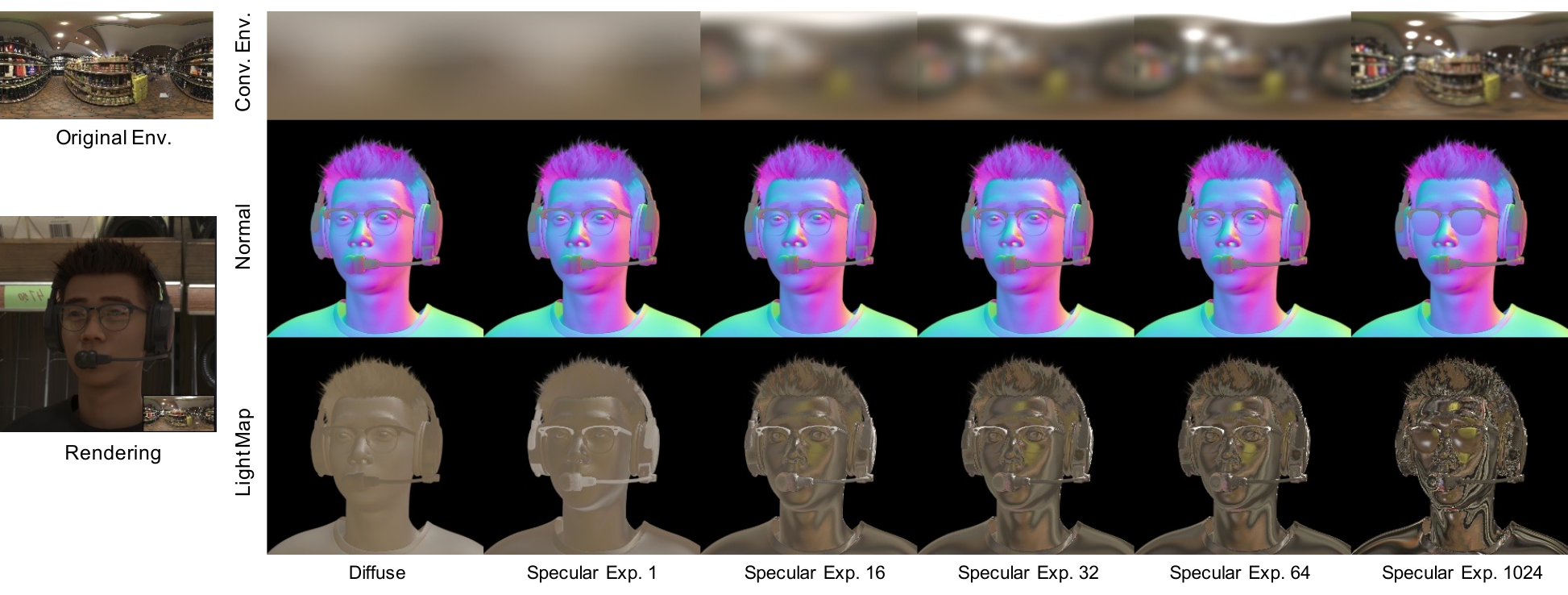}
    \vspace{-0.7cm}
  \caption{Example of light maps. We compute prefiltered environment maps by diffuse and specular convolution. We then compute diffuse light map $L_d$ and specular light maps $L_s^1$, $L_s^{16}$, $L_s^{32}$, $L_s^{64}$ by indexing the prefiltered maps using normal without glasses $N$ and using normal with glasses $N_l$ to compute ${L}_{s,l}^{1024}$.}
  \vspace{-0.3cm}
\label{fig:light_map}
\end{figure*}
In addition to the rendering example shown in Figure~\ref{fig:rendering}, our synthetic data generation pipeline can generate other intrinsic components rendered but unused in our framework. As shown in Figure~\ref{fig:rendering_more}, we can also render diffuse, specular, subsurface scattering, coat, and sheen components which all rendered from the Arnold \textit{aiStandardSurface} shader. We will release our synthetic dataset.

As shown in Figure~\ref{fig:shader_map}, we use a mask in the texture space to control different parameters for face and eye regions. For face regions, we apply 0.95 weight to subsurface scattering (SSS) and set the RGB texture (diffuse albedo) as SSS color, (1, 0.35, 0.2) as RGB Radius, 0.225 as scale, \textit{randomwalk\_v2} as type, and 0 as anisotropy. We apply a specular albedo map to specular weight. Then, we set 0.82 to specular color and 0.4 to specular roughness, respectively. We set 0.1 to coat weight, a micro-specular albedo map (UV repeated 160 times) to coat color, 0.25 to coat roughness, and 1.5 as coat IOR. We set 0.1 as sheen weight, 1.0 as sheen color, and 0.3 as sheen roughness. We apply a micro-normal map (UV repeated 160 times) as a bump map. We apply a displacement map (computed from a raw 3D face model) to skin regions. 

The eye and inner mouth regions of our models are represented by a billboard mesh that connects upper and lower eyelids or lips vertices with a projected texture similar to~\cite{Lombardi:2018}. 

For eye regions, we set the RGB texture (diffuse albedo) as the base color. We apply a gradient map (0.2 at the center, 1.0 at the boundary, computed from $1-0.8\cdot$ \textit{eye displacement} below) to two eyeball regions of the diffuse texture to compensate for the unwanted refection baked from the data capture. Specular weight and color are the same as the face, but we set 0.1 for specular roughness. We set 0.5 to coat weight, the same micro-specular albedo map for coat color, 0.1 for coat roughness, and 1.376 for coat IOR. The same micro-normal map is applied as a bump map. We also set a gradient map as the displacement map to simulate the eyeball geometry which is otherwise flat in the original face model. There are no SSS and sheen applied to eye regions. In addition to the components from the shader, we also render shadow maps and object ID maps, which might be useful for future research.

We also demonstrate the light maps computed from the input HDR environment map and normal in Figure~\ref{fig:light_map}.

\subsection{User study interface}
\label{sec_a:user_study}
\begin{figure*}[t]
\centering
\includegraphics[width=\linewidth]{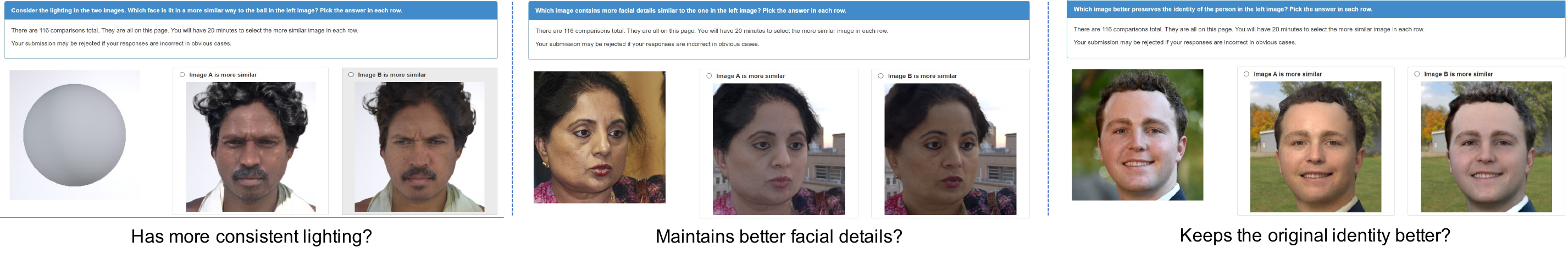}
    \vspace{-0.7cm}
  \caption{Examples of user study questions interface for the three questions: \textit{1) has more consistent lighting? 2) maintains better facial details? 3) keeps the original identity better?} For the first question, we present a diffuse ball lit by the target environment map and ask users to select an option from two relit images, which is lit in a more similar way to the ball. For the last two questions, we present an input portrait image and ask users to select an option on relit results from two methods. }
\label{fig:user_study}
\end{figure*}
We demonstrate our user study interface for the questions: given an input image and an environment map, which relit image \textit{1) keeps the original identity better? 2) maintains better facial details? 3) has more consistent lighting with the environment map?} between our relighting result and one of the baselines in Figure~\ref{fig:user_study}. For the first two questions, we present an input portrait image and ask users to select an option on relit results from two methods. For the last question, we present a diffuse ball that is lit by the target environment map and ask users to select an option from two relit images which is lit in a more similar way to the ball.

\subsection{OLAT details}
\label{sec_a:olat}
\begin{figure*}[t]
\centering
\includegraphics[width=0.98\linewidth]{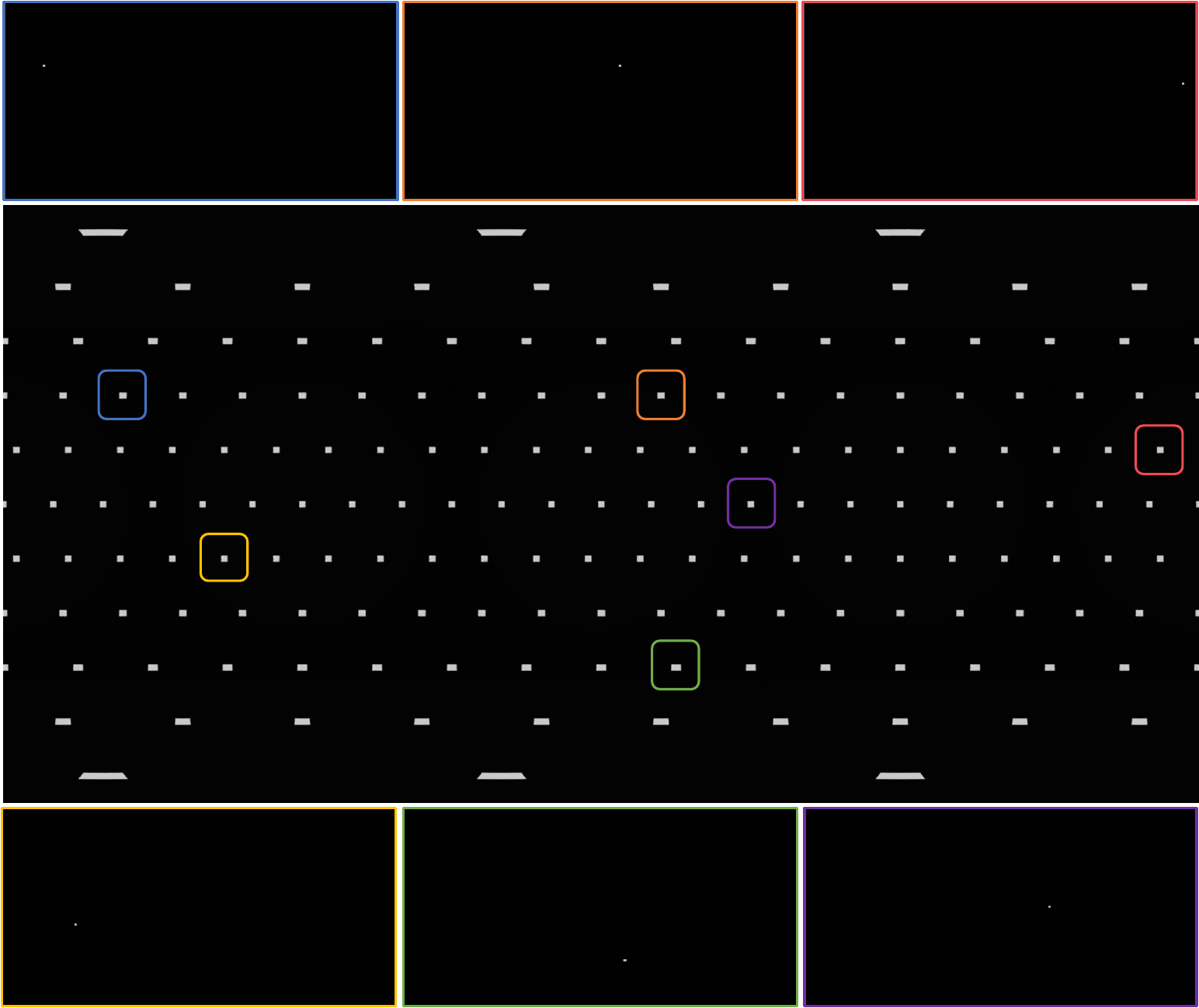}
    \vspace{-0.3cm}
  \caption{We created OLAT-like HDR environment maps at uniformly sampled locations on a sphere. In the middle we show all the sampled locations (all lights on). In the top and bottom row, we show different OLAT environment maps (one light on) at 6 locations.}
  \vspace{-0.3cm}
\label{fig:olat}
\end{figure*}
We demonstrate the one-light-at-a-time (OLAT) images we used to train and evaluate our network in Figure~\ref{fig:olat}. In total, we uniformly sample 168 locations on a sphere. We put a high-intensity light source at each location and project them to latitude-longitude format to create 168 OLAT HDR environment maps in total. Each light source in our OLAT HDR environment is created such that the light source preserves the same intensity when it is mapped on to the sphere from the latitude-longitude representation.

\end{document}